\documentclass[a4paper,fleqn]{cas-dc}

\usepackage[authoryear]{natbib}
\usepackage{booktabs}
\usepackage{algorithmic}
\usepackage{algorithm}
\usepackage{array}
\usepackage{subcaption}
\usepackage{textcomp}
\usepackage{stfloats}
\usepackage{url}
\usepackage{verbatim}
\usepackage{graphicx}
\usepackage{adjustbox}
\usepackage{multirow}
\usepackage{pifont}
\usepackage{colortbl}
\usepackage{threeparttable}
\usepackage{makecell}
\usepackage{hyperref}
\usepackage{stfloats}
\usepackage{placeins}
\def\tsc#1{\csdef{#1}{\textsc{\lowercase{#1}}\xspace}}
\tsc{WGM}
\tsc{QE}
\begin{document}
\let\WriteBookmarks\relax
\def\floatpagepagefraction{1}
\def\textpagefraction{.001}

\shorttitle{Accident Analysis and Prevention}    

\shortauthors{Ronghui Zhang et~al.}  

\title [mode = title]{Lane Departure Accident Prevention in Foggy Conditions: A Prior-Guided Dynamic Feature Fusion Transformer Framework for Real-Time Lane Detection}


% \ead{}


\author[1]{Ronghui Zhang}
\credit{Conceptualization, Methodology, Data curation, Investigation, Writing–original draft, Writing–review and editing, Software, Validation, Funding acquisition, Project administration, Supervision, Resources}
\author[1]{Yuhang Ma}
\credit{Conceptualization, Methodology, Data curation, Investigation, Writing–original draft}
\author[1]{Tengfei Li}
\credit{Conceptualization, Methodology, Data curation, Investigation, Writing–original draft}
\author[1]{Ziyu Lin}
\credit{Conceptualization, Methodology, Data curation, Investigation, Writing–original draft}
\author[1]{Xiao Li}
\credit{Conceptualization, Methodology, Data curation, Investigation, Writing–original draft}
\author[1]{Yueying Wu}
\credit{Conceptualization, Methodology, Data curation, Investigation}
\author[1]{Junzhou Chen}[orcid=0000-0002-3388-3503]
\credit{Data curation, Investigation, Writing–review and editing, Funding acquisition, Project administration, Supervision, Resources}
\cormark[1]
\author[2]{Qiang Zeng}
\credit{Writing–review and editing, Project administration, Supervision, Resources}
\author[3]{Lin Zhang}
\credit{Writing–review and editing, Project administration, Supervision, Resources}
\author[4]{Jia Hu}
\credit{Writing–review and editing, Project administration, Supervision, Resources}
\author[5]{Tony Z. Qiu}
\credit{Writing–review and editing, Project administration, Supervision, Resources}
\author[6]{Konghui Guo}
\credit{Writing–review and editing, Project administration, Supervision, Resources}

\affiliation[1]{organization={Guangdong Key Laboratory of Intelligent Transportation System, School of Intelligent Systems Engineering, Sun Yat-sen University},
             city={Guangzhou},
             postcode={510275},
             state={Guangdong},
             country={China}}

\affiliation[2]{organization={School of Civil Engineering and Transportation, South China University of technology},
            city={Guangzhou},
            postcode={510640}, 
            state = {Guangdong},
            country={China}}
                    
\affiliation[3]{organization={College of Automotive Studies, Tongji University},
             city={Shanghai},
             postcode={201804},
             country={China}}

\affiliation[4]{organization={College of Transportation Engineering, Tongji University},
             city={Shanghai},
             postcode={201804},
             country={China}}

\affiliation[5]{organization={Department of Civil and Environmental Engineering, University of Alberta},
             city={Edmonton},
             %postcode={},
             state={Alberta},
             country={Canada}}

\affiliation[6]{organization={State Key Laboratory of Automotive Chassis Integration and Bionics, Jilin University},
             city={Changchun},
             postcode={130025},
             state={Jilin},
             country={China}}

\cortext[1]{Corresponding author}



\begin{abstract}
Lane departure accident prevention plays a critical role in enhancing road safety, and lane detection is a core technology to achieve this goal, especially under complex weather conditions. While existing lane detection algorithms perform well under favorable weather conditions, their effectiveness significantly degrades in foggy environments, which increases the risk of traffic accidents. In response to this challenge, we propose PDT-Net, a robust Prior-Guided Dynamic Feature Fusion Transformer framework designed for real-time lane detection in foggy conditions. This framework integrates three key modules: a Global Feature Fusion Module (GFFM) to capture the relationship between local and global features in foggy images, a Dynamic Feature Fusion Module (DFFM) to model the structural and positional relationships of lane instances, and a Prior-Guided Edge Enhancement Module (PEM) to recover lost edge details in foggy environments. Furthermore, we introduce the FoggyLane dataset, a real-world dataset that specifically targets lane detection in foggy conditions, along with two synthesized datasets, FoggyCULane and FoggyTusimple, to address the lack of fog-specific data for lane detection. Extensive experiments show that PDT-Net achieves state-of-the-art performance with F1-scores of 95.04\% on FoggyLane, 79.85\% on FoggyCULane, and 96.95\% on FoggyTusimple. Moreover, with TensorRT acceleration, our method achieves a processing speed of 38.4 FPS on the NVIDIA Jetson AGX Orin, confirming its real-time capability and robustness in challenging foggy environments. By improving the precision of lane detection, our framework can contribute to active safety warning systems, helping to prevent accidents in foggy conditions.
\end{abstract}

% \begin{highlights}
% \item 
% \item 
% \item 
% \end{highlights}


\maketitle

% Main text
\section{Introduction}
Road traffic crashes continue to be a major global safety concern, with lane departure being a significant contributor. According to the National Highway Traffic Safety Administration (NHTSA), over 40\% of fatal accidents in 2023 were attributed to lane departure\citep{NHTSA2025}, primarily caused by adverse weather conditions, degraded road markings, and driver errors. These factors expose critical gaps in vehicle perception, highlighting the need for more robust lane detection systems to address this issue.

Computer vision techniques, particularly those focused on real-time risk assessment and accident prediction, play a pivotal role in enhancing active safety warning systems\citep{AAP4}. Lane detection, as a core perceptual capability, enables Advanced Driver Assistance Systems (ADAS) to trigger active safety warnings such as lane departure warning, which are essential for accident prevention.

\begin{figure}
    \centering
    \includegraphics[width=\columnwidth]{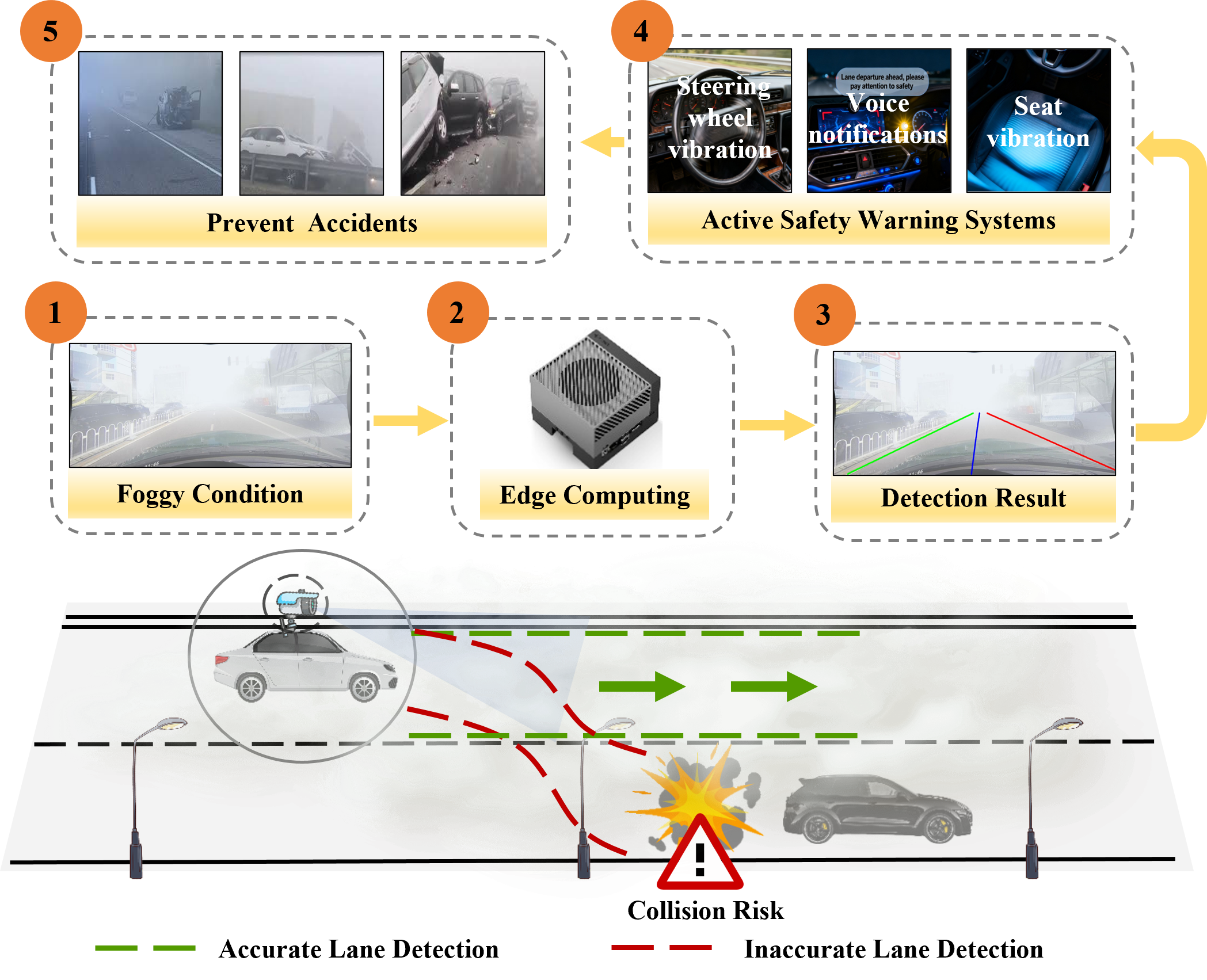}
    \caption{Lane detection in foggy scenarios for ADAS.\citep{Pexels} Inaccurate lane detection may lead to lane departure, increasing the risk of collision. Accurate lane detection provides essential technical support for the active safety functions of vehicles. Once the vehicle deviates from its intended lane, the ADAS can promptly issue an active safety warning to alert the driver, enabling timely corrective actions and ultimately preventing the occurrence of traffic accidents.}
    \label{fig_Scene}
\end{figure}

Under good weather and clear lane markings, current lane detection algorithms perform well. However, in complex road environments and poor weather, such as those with poor weather conditions like fog, lane markings can become blurred, image quality may decrease, and global environmental information is often obscured, which makes lane detection more challenging. In these conditions, existing algorithms often struggle and may fail, leading to an increased risk of traffic accidents with more severe consequences. Ensuring safe driving in these conditions calls for the development of high-precision, highly robust lane detection methods optimized for foggy environments. As shown in Fig. \ref{fig_Scene}, The in-vehicle camera captures real-time road images, which are then preprocessed by edge devices and sent to the lane detection module. The detected lane information is used for active safety alerts. In the event of lane departure, the system proactively issues warnings\citep{AAP7} through steering wheel vibration, voice notifications, and seat vibration, helping to prevent accidents\citep{AAP8}.

Earlier lane detection algorithms\citep{son2015real, hou2016efficient} primarily relied on traditional computer vision methods, such as feature-based approaches that detect vanishing points, use color features, and apply edge detection or line detection techniques like Hough transforms. These methods are efficient for real-time applications but require manual parameter tuning and often struggle in challenging scenarios like worn or occluded lane markings, which may lead to serious collision accidents.

With the rise of deep learning, many modern lane detection methods \citep{neven2018towards, yu2018deep, pan2018spatial, abualsaud2021laneaf} leverage CNNs, which automatically extract features from raw data and generalize better in complex road environments. While deep learning provides higher robustness, many models still rely on pixel-by-pixel classification, which is computationally expensive and unnecessary for lane detection. Moreover, lane detection doesn’t always require dense pixel segmentation, which can be inefficient.

Additionally, some approaches have incorporated anchor-based methods from object detection to improve efficiency and accuracy, particularly for complex or occluded lanes. These methods preset anchors at different starting points and slopes to optimize lane shape and position prediction \citep{AAP1, redmon2016you, farhadi2018yolov3, chen2019pointlanenet, zheng2022clrnet, AAP5}. However, these techniques still face challenges in handling foggy conditions, where image degradation severely impacts lane visibility. Traditional image enhancement methods, like dehazing, are commonly used to improve clarity but do not always result in better detection performance \citep{huang2020dsnet}. Some approaches generate synthetic foggy images for data augmentation, but these do not always capture real-world complexities \citep{nie2022sensor}.

Moreover, domain adaptation techniques \citep{chen2018domain, zhang2021domain, hnewa2021multiscale, li2023domain} have been employed to bridge the gap between foggy and clear-weather images. These methods often rely on labeled clear-weather datasets or use transfer learning, but they do not fully address the unique challenges posed by foggy conditions.

Lane detection enables key functionalities such as lane departure warnings and autonomous navigation in ADAS, demonstrating strong risk perception capabilities\citep{AAP3, AAP6}. While existing lane detection algorithms perform well under clear weather conditions with visible lane markings, their performance drops significantly in adverse environments like fog, where reduced visibility and image degradation obscure lane clarity, thus increasing the risk of traffic accidents. Furthermore, the lack of specialized datasets tailored to foggy conditions has limited the development of effective deep learning-based solutions for these challenging scenarios.

To address these limitations, we introduce the FoggyLane dataset, a new benchmark specifically designed for lane detection under foggy conditions. This dataset, comprising 1,423 annotated images across six distinct foggy road scenarios, fills a crucial gap in the current lane detection datasets. Additionally, we create foggy versions of two commonly used lane detection datasets, CULane \citep{pan2018spatial} and Tusimple \citep{Tusimple}, resulting in the FoggyCULane and FoggyTusimple datasets. These datasets are essential for training and testing lane detection algorithms in foggy environments.

In response to the unique challenges posed by foggy conditions, we propose a robust real-time lane detection method optimized for such environments. The main contributions of our work are as follows:

\begin{itemize}
\item [1)] To address the issue of missing global information in foggy images, we design a Global Feature Fusion Module (GFFM) within the backbone network to capture relationships between inputs, improving feature extraction in foggy scenarios.
\item [2)] To enhance the utilization of structural relationships between lane instances, we introduce a Dynamic Feature Fusion Module (DFFM) that learns and predicts the correlations between lane instances, improving lane detection accuracy and robustness.
\item [3)] To tackle the loss of edge information in foggy images, we incorporate a Prior-Guided Edge Enhancement Module (PEM), which enhances the model's sensitivity to fine edge details. This ensures better lane delineation, even when visibility is severely impaired by fog, and significantly improves the reliability of lane departure warnings.
\item [4)] To address the lack of open-source foggy lane detection datasets, we construct FoggyLane, a real-world dataset with 1,423 annotated images, and generate FoggyCULane and FoggyTusimple based on existing datasets. Our method achieves state-of-the-art performance with inference speeds of 192.9 FPS on NVIDIA RTX 3090 and 38.4 FPS on NVIDIA Jetson AGX Orin, confirming its real-time capability and practical applicability in real-world settings.
\end{itemize}

\section{Related Work}
In recent years, advances in deep learning have substantially improved lane detection performance. Building on these advances, current lane detection methods are divided into four main paradigms: segmentation-based, anchor-based, curve-based, and row-wise-based approaches, according to their modeling strategies.

\subsection{Segmentation-based Lane Detection}
Segmentation-based methods mainly utilize semantic segmentation or instance segmentation techniques to distinguish lane lines from other objects or backgrounds in the image, transforming lane detection into a pixel-level classification problem. LaneNet\citep{neven2018towards} employs a two-branch structure of split branches and embedded branches, the split branches output a binary lane mask to recognize lane lines, and the embedded branches assign a unique embedding to each pixel to realize the distinction of lane lines. SCNN\citep{pan2018spatial} modifies the conventional layer-wise convolution approach by incorporating a slice-wise convolution technique, enabling horizontal and vertical information flow between pixels within a layer. RESA\citep{zheng2021resa} introduces a spatial attention module that ensures robust lane prediction by cyclically shifting the feature map horizontally and vertically so that each pixel can acquire global information. Although these segmentation-based methods possess high accuracy and robustness, the models are bulky and slow to process due to high consumption of computational resources. HW-Transformer\citep{tits3} presents an innovative lane detection network to expand the visual range around the lane and enable global information communication through intersecting features, alongside a self-attention knowledge distillation (SAKD) method to enhance performance and semantic feature learning.

\subsection{Curve-based Lane Detection}
Curve-based methods transform the lane detection into a regression task by predicting polynomial coefficients that fit lane lines to polynomial equations. PolyLaneNet\citep{tabelini2021polylanenet} proposes an end-to-end convolutional neural network that transforms images into polynomials representing each lane's markers, along with domain-specific lane polynomials and confidence values for every lane. Similarly, LSTR\citep{liu2021end} utilizes a Transformer-based architecture that captures both local lane structures and broader contextual information through non-local interactions. DBNet\citep{dai2024dbnet} introduces NURBS curves and utilizes their geometric semantic properties to achieve local and global optimization, which improves the robustness of lane line detection and model interpretation. However, polynomial fitting is sensitive to parameter selection, and minor changes in parameters can result in significant variations in fitting results, leading to unstable lane line detection. 
\subsection{Anchor-based Lane Detection}
Anchor-based approaches employ predetermined reference lines within the image space to generate lane predictions, subsequently calculating necessary offsets to precisely match these anchors with true lane markings. Line-CNN\citep{li2019line} introduces the anchor mechanism into lane detection for the first time, and innovatively proposes line anchors. LaneATT\citep{tabelini2021keep} effectively solves the problem of occlusion and illumination due to the lack of a line anchor, by fusing extracted anchor features with global features generated by attention module. CLRNet\citep{zheng2022clrnet} employs a coarse-to-fine detection strategy, where lane lines are first approximated using high-level semantic cues before being refined with detailed low-level features. This hierarchical fusion of semantic information significantly improves detection precision. FLAMNet\citep{ran2023flamnet} employs an adaptive line anchor strategy, augmented by patch-based feature pooling and decomposed self-attention mechanisms. This design strengthens both local feature extraction and global context modeling, improving performance in handling intricate lane structures. However, due to the multiple curvature and orientation variations of lane lines in the real world, the performance of the anchor-based methods may be limited in capturing precise paths with complex shapes. 

\subsection{Row-wise-based Lane Detection}
Row-wise methods predict lane positions for each image row. E2E-LMD\citep{yoo2020end} introduces the row-wise-based lane detection task and develops a complete architecture that outputs lane marker positions directly. UFLD\citep{qin2020ultra} utilizes global features to select the lane positions of predefined rows of an image and proposes a structural loss function, so as to ensure the continuity and smoothness of the lanes. CondLaneNet\citep{liu2021condlanenet} utilizes conditional convolution and row-by-row prediction strategies to enhance lane recognition accuracy, and introduces the RIM module to effectively address complex lane layouts. The row-wise-based detection methods improve computational efficiency by focusing only on the row regions in the image, and possesses high detection accuracy while maintaining efficiency.

\begin{figure*}[htb]
    \centering
    \includegraphics[width=\textwidth]{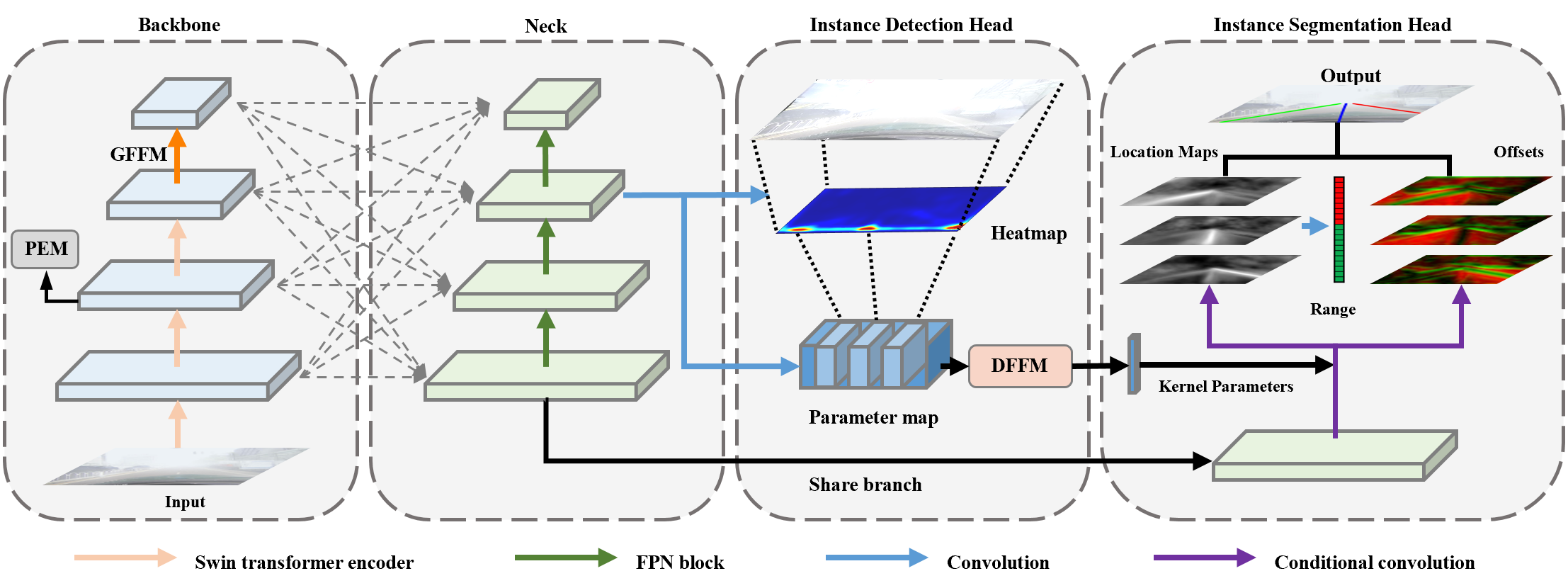}
    \caption{Overview of the proposed network.}
    \label{fig_framework}
\end{figure*}

\begin{figure*}[htb]
    \centering
    \includegraphics[width=\textwidth]{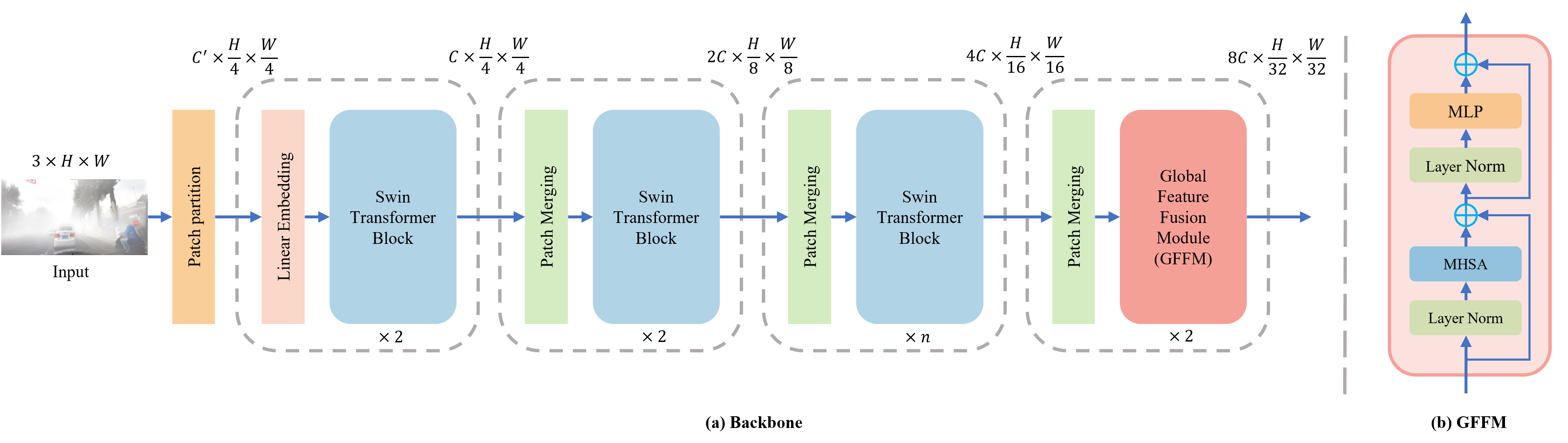}
    \caption{Backbone Network with Global Feature Fusion Module(GFFM).}
    \label{fig_GFFM}
\end{figure*}

\section{Methods}
\subsection{Overall structure}
To address the challenge of lane detection in foggy conditions, where visibility is significantly reduced, we propose PDT-Net, a robust Prior-Guided Dynamic Feature Fusion Transformer framework. The network works through several collaborative modules to tackle the challenges posed by foggy images, with the overall architecture shown in Fig. \ref{fig_framework}. The input image first passes through the backbone network, where feature extraction occurs, followed by the Global Feature Fusion Module (GFFM). This module is designed to mitigate the loss of global information in foggy images, allowing the network to retain essential global context. The output is a feature map enriched with both local and global information, crucial for accurate lane detection.

The multi-level feature map from the backbone is then passed into the Neck section of the network, which employs HRNetV2\citep{9052469} for multi-scale feature fusion. HRNetV2 excels in integrating high-resolution and low-resolution features from different stages of the backbone. This process helps the network capture spatial information at multiple scales, producing a rich, multi-scale feature representation that forms the basis for lane instance detection.

Next, the Instance Detection Head processes the multi-scale feature map from the Neck network, predicting dynamic convolution kernel parameters for each lane instance. These dynamic kernels are applied to the lane mask segmentation head to isolate instance-specific features. The output includes a parameter map and a heat map, crucial for identifying lane instances. This approach eliminates the need for complex post-processing, significantly enhancing detection speed and accuracy.

The output parameter map is then passed to the Dynamic Feature Fusion Module (DFFM), which learns the relationships between different lane instances. The DFFM fuses these instance-specific features, improving lane localization accuracy, especially for complex or overlapping lane lines. This module ensures that the network can better identify lane instances and refine the localization of each lane.

Finally, to compensate for the loss of edge information in foggy images, the network incorporates a Prior-Guided Edge Enhancement Module (PEM). This module enhances edge features using convolution operations, recovering fine details at lane boundaries. The enhanced edge features improve the network's ability to detect lane boundaries with high precision, even under challenging foggy conditions.

\subsection{Global Feature Fusion Module}
As shown in Fig. \ref{fig_GFFM}, to enhance global feature integration in the Swin Transformer backbone\citep{liu2021swin}, we replace the final window-based attention mechanism with a standard multi-head attention mechanism (MHSA) from Vision Transformer (ViT)\citep{vit}. While ViT excels at capturing global dependencies, it suffers from high computational cost, especially at high resolutions, making it unsuitable for real-time tasks such as lane detection. The traditional ViT computes global self-attention with a complexity of \( O(N^2) \), where \( N \) is the number of input tokens. This high computational cost becomes prohibitive at high resolution, slowing down inference and limiting real-time processing.

In our approach, we address this issue by introducing an efficient solution: global attention is applied only after reducing the image resolution. This ensures that the computational cost of global attention is manageable\citep{AAP2}, allowing us to retain the global attention mechanism of ViT while maintaining fast inference speed, which is crucial for real-time lane detection in foggy conditions.

\begin{figure*}[b]
    \centering
    \includegraphics[width=\textwidth]{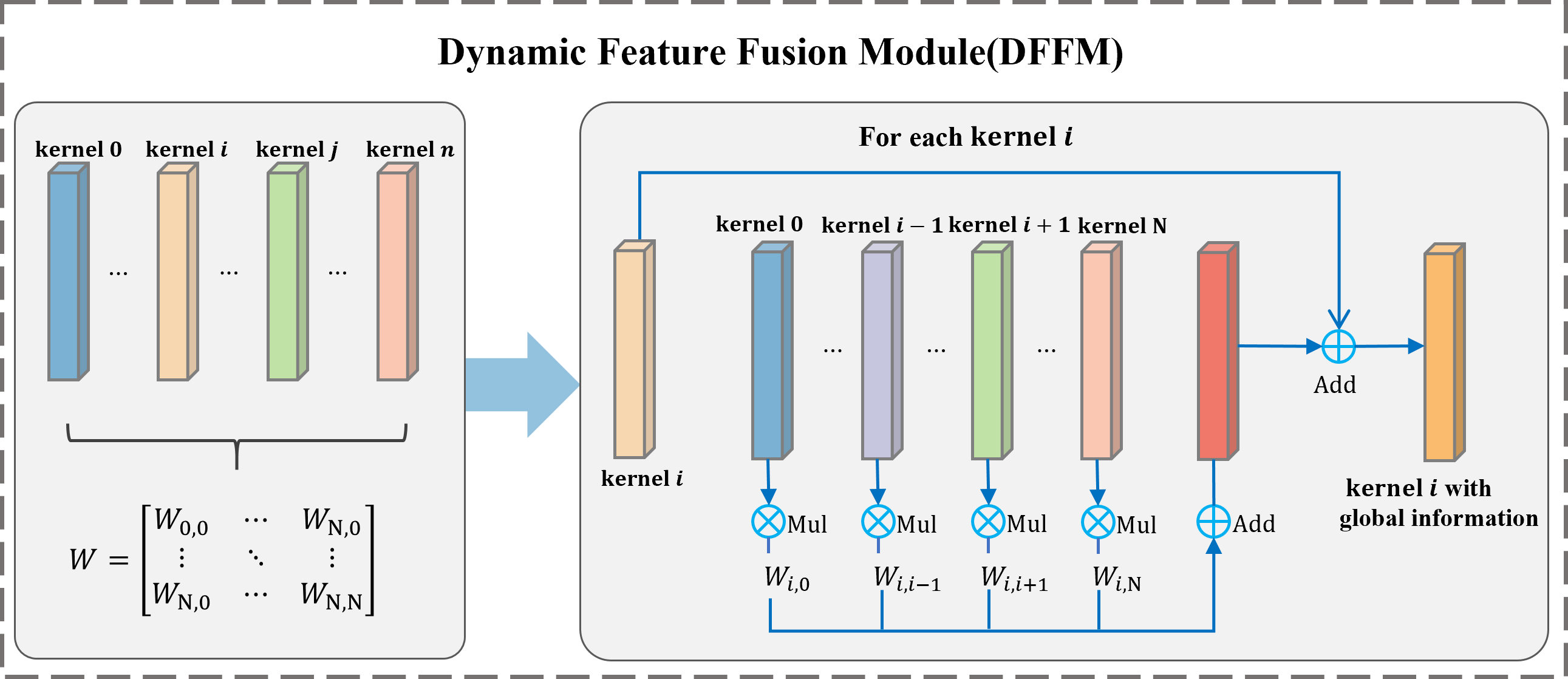}
    \caption{The structure of the Dynamic Feature Fusion Module(DFFM)}
    \label{fig_DFFM}
\end{figure*}

In our design, this MHSA layer replaces the original window-based attention mechanism used in the Swin Transformer. By applying global attention after reducing the image resolution, we strike a balance between maintaining global context awareness and achieving high computational efficiency, which is crucial for real-time lane detection tasks.

\subsection{Dynamic Feature Fusion Module}
Existing lane instance segmentation methods\citep{tian2020conditional, liu2021condlanenet}, typically use dynamic convolution to generate kernel parameters for each instance, which improves the network's flexibility and adaptability. However, these methods do not fully leverage the structural relationships between lane instances, leading to limited segmentation and detection accuracy. Traditional dynamic convolution methods only consider individual lane instances and overlook the potential relationships between adjacent instances, which reduces their performance when handling complex lane scenes.
To address this limitation, we propose the Dynamic Feature Fusion Module (DFFM), as shown in Fig. \ref{fig_DFFM}, which optimizes kernel generation by integrating information from adjacent lane instances. Specifically, DFFM introduces structured feature fusion into dynamic convolution, allowing each lane instance's kernel to depend not only on the features of the current lane but also on the information from nearby lane instances. This approach enables the network to adaptively learn the relationships between lane instances, improving both detection accuracy and segmentation performance.

Let the input feature map from the lane instance detection branch be denoted as \( \mathit{K} = \left[ \mathit{kernel}_0, \mathit{kernel}_1, \dots, \mathit{kernel}_N \right] \), where each \( \mathit{kernel}_i \in \mathbb{R}^{k \times k \times C} \), and define the weight matrix \( \mathit{W} \in \mathbb{R}^{(N+1) \times (N+1)} \) to capture the contributions of each kernel to the final fused kernel.

The final fused kernel for lane \( i \) is computed as the weighted sum of the kernels:
\begin{equation}
\hat{\mathit{kernel}}_i = \sum_{j=0}^{N} W_{i,j} \cdot \mathit{kernel}_j
\end{equation}
where \( W_{i,j} \) is the weight for the contribution of \( \mathit{kernel}_j \) to the fused kernel \( \hat{\mathit{kernel}}_i \); The weight matrix \( \mathit{W} \) is learned during training.

The final fused kernel is obtained through matrix multiplication as follows:
\begin{equation}
\hat{\mathit{K}}_i = \mathit{W}_i \cdot \mathit{K}
\end{equation}
where \( \mathit{W}_i \) corresponds to the row of the weight matrix \( \mathit{W} \) associated with kernel \( i \).

The resulting fused kernel \( \hat{\mathit{kernel}}_i \) is then used in the dynamic convolution operation for lane segmentation, leading to more accurate lane detection and segmentation.

\subsection{Prior-Guided Edge Enhancement Module}
\label{sec:edge-enhancement}
\begin{figure}
    \centering
    \includegraphics[width=\columnwidth]{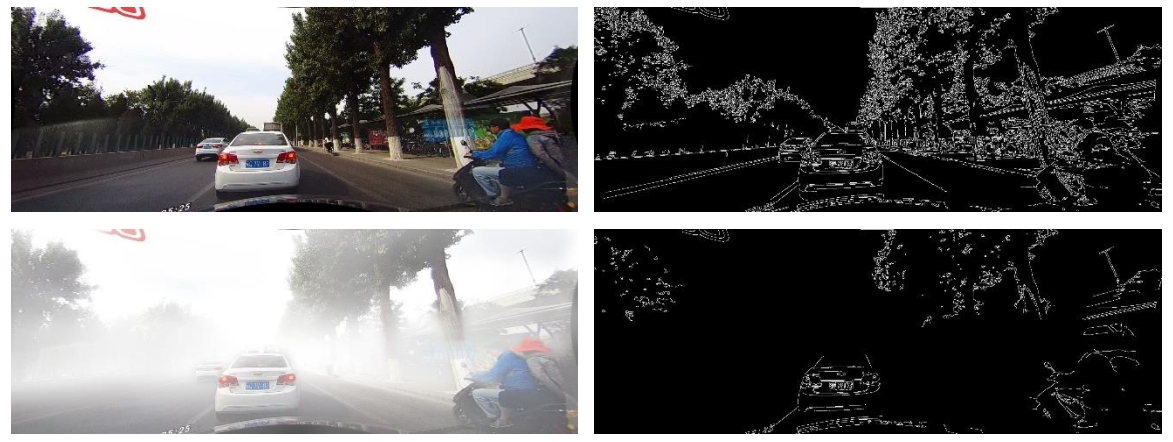}
    \caption{Comparison of edge information between foggy and normal scenarios.}
    \label{fig_edge_information}
\end{figure}

\begin{figure*}[!htb]
    \centering
    \includegraphics[width=\textwidth]{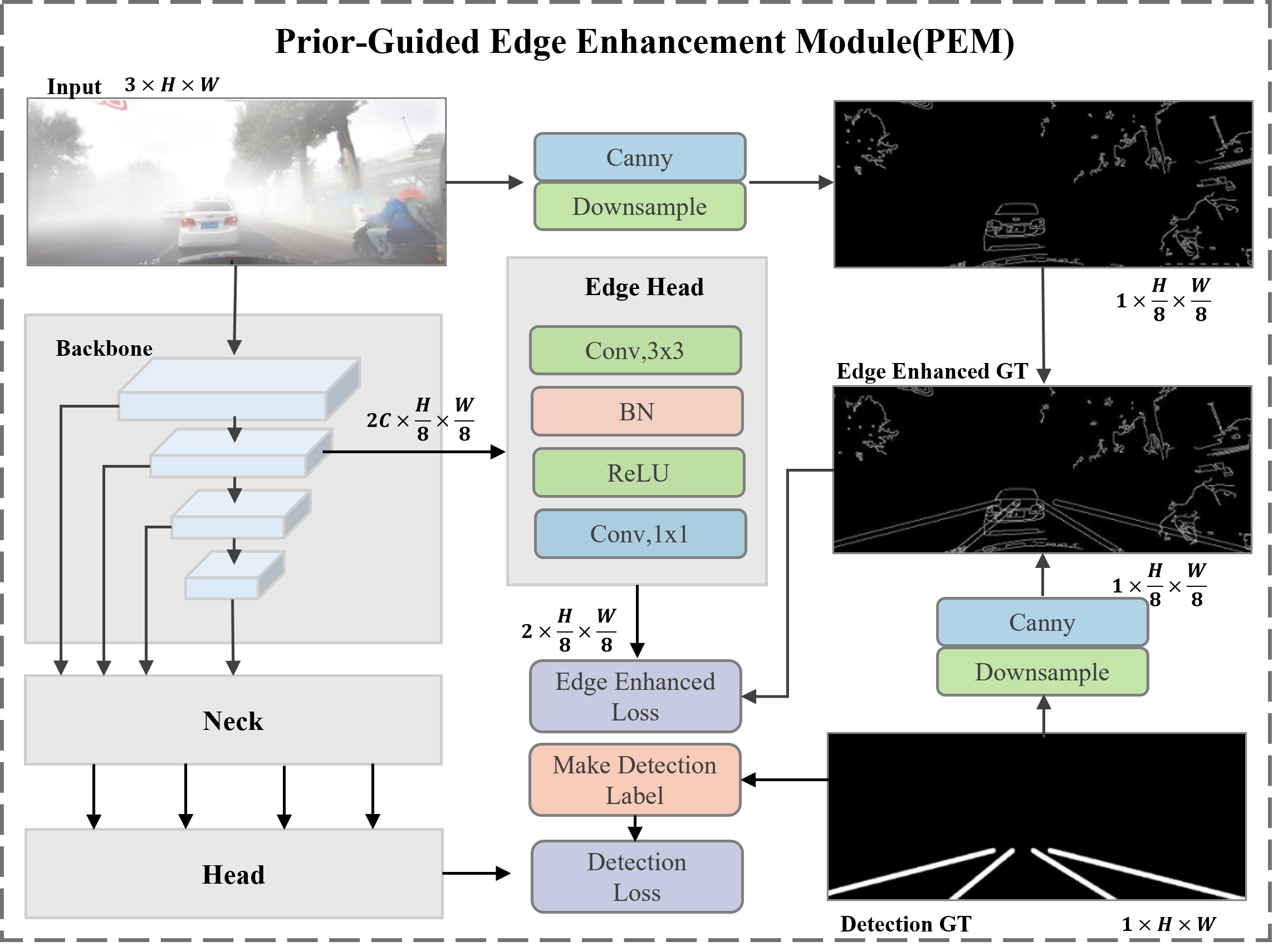}
    \caption{The structure of the Prior-Guided Edge Enhancement Module(PEM)}
    \label{fig_PEM}
\end{figure*}

In foggy conditions, images are typically characterized by reduced contrast and sharpness, making it difficult to detect key features such as edges\ref{fig_edge_information}, which are critical for tasks like lane detection. In these scenarios, edge information becomes an essential prior knowledge that can guide the network in recovering vital details. Edge information not only defines object boundaries but also conveys structural and spatial relationships, crucial for understanding the scene. Recognizing this, we propose the Prior-Guided Edge Enhancement Module (PEM),as shown in Fig. \ref{fig_PEM}, which integrates prior knowledge of edge structures to recover lost edge information in foggy images, ultimately enhancing lane detection accuracy and robustness.

PEM works by adding an auxiliary path to the network that specifically learns edge features of the input image. The process begins with the application of the Canny operator for edge detection, followed by the reconstruction of missing lane edges using lane line labels. This ground truth for the edge enhancement module is generated by combining detected edges with labeled edges. This approach helps the network better identify lane boundaries, which are often difficult to distinguish in foggy conditions.

In the second stage of the backbone network, we introduce a branch that processes the extracted edge features. This branch sequentially applies a \( 3 \times 3 \) convolution followed by a \( 1 \times 1 \) convolution to enhance edge-related features. The mathematical representation of this process is as follows:
\begin{equation}
\hat{\mathit{E}}_{\mathit{predict}} = \mathit{Conv}_2\left(\mathit{ReLU}\left(\mathit{BN}(\mathit{Conv}_1(\mathit{X}))\right)\right)
\end{equation}
where \( \mathit{Conv}_1 \) is a \( 3 \times 3 \) convolution designed to capture spatial edge features, \( \mathit{BN} \) is the batch normalization layer to stabilize the learning process by normalizing the output, \( \mathit{ReLU} \) is the activation function that provides non-linearity, \( \mathit{Conv}_2 \) is a \( 1 \times 1 \) convolution used to reduce the dimensionality and refine edge features.

During training, the edge information extracted via the Canny operator from both the input image and ground truth (GT) lane lines is first downsampled to match the resolution of the network's predicted edge information \( \hat{\mathit{E}}_{\mathit{predict}} \). Then, these two edge information maps are merged. The final combined edge information is represented as \( \hat{\mathit{E}}_{\mathit{GT}} \), which can be computed as:
\begin{equation}
\hat{\mathit{E}}_{\mathit{GT}} = \mathit{Edge}_I + \mathit{Edge}_{\mathit{Lane}}
\end{equation}
where \( \mathit{Edge}_I \) is the edge information obtained from the input image \( \mathit{I} \) using the Canny operator, \( \mathit{Edge}_{\mathit{Lane}} \) is the edge information derived from the lane ground truth (GT) using the Canny operator.

Next, we compute the cross-entropy loss between the predicted edge information \( \hat{\mathit{E}}_{\mathit{predict}} \) and the final combined edge information \( \hat{\mathit{E}}_{\mathit{GT}} \). This loss function is formulated as:
\begin{equation}
L_{\mathit{edge}} = - \sum_{i} \left( e_{\mathit{gt}}^i \log(p_i) + (1 - e_{\mathit{gt}}^i) \log(1 - p_i) \right)
\end{equation}
where \( e_{\mathit{gt}}^i \) represents the ground truth label for pixel \( i \) (1 if it’s part of the edge, 0 otherwise), \( p_i \) is the predicted probability that pixel \( i \) is part of the edge.

The PEM module allows the network to effectively enhance critical edge details in lane detection, significantly improving both the accuracy and stability of lane detection in foggy conditions. This module not only recovers lost edge information in images but also integrates information from various sources, using the cross-entropy loss function to optimize the network. As a result, PEM plays a key role in enhancing the robustness of lane detection systems in challenging environments.

\subsection{Loss functions}
Based on the above, our method involves a total of five prediction tasks: lane start point prediction, row-based segmentation prediction, vertical range prediction, offset prediction, and edge prediction. The loss functions for each task are detailed below.
\subsubsection{Lane start point prediction}
In feature maps with a large receptive field and low resolution, the lane line start point is difficult to be accurately represented by a single pixel location. Moreover, in such large receptive field feature maps, pixels near the target point exhibit similar features. If these pixels are directly labeled as negative samples, it will interfere with the network training and make it difficult for the network to learn and optimize. Therefore, this method uses a Gaussian kernel function to generate a heatmap label for the lane line start point. Each element in the heatmap label represents the confidence that the position corresponds to the target point; the closer to the target point, the higher the confidence, and the farther away, the lower the confidence. In other words, the ground truth is softly labeled with a Gaussian function, making it easier for the network to converge.
Suppose \((x, y)\) denotes a position on the low-resolution feature map, and \((p_x, p_y)\) represents the coordinate of the target point on the low-resolution feature map. Then, the ground truth heatmap \(GT\) value at position \((x, y)\) is defined as:
\begin{equation}
GT(x,y) = \exp\left(-\frac{(x - p_x)^2 + (y - p_y)^2}{2\sigma^2}\right)
\end{equation}
where \(\sigma\) controls the decay rate of the Gaussian kernel. A larger \(\sigma\) results in slower confidence decay and thus a larger heatmap area.

This method combines the Gaussian heatmaps of all lane line start points within the image into a single heatmap. When multiple heatmaps respond at the same location, the maximum value is retained to form the final ground truth heatmap. Then, the Focal Loss is used to calculate the loss between the predicted heatmap and the ground truth heatmap. Let the predicted heatmap value at position \((x,y)\) be \(P_{xy}\), and let \(N\) be the total number of pixels in the heatmap. The Focal Loss between the predicted heatmap and the ground truth heatmap is defined as:

\begin{equation}
\resizebox{1\linewidth}{!}{$
\mathcal{L}_{\text{hm}} = -\frac{1}{N} \sum_{x,y} \begin{cases}
(1 - P_{xy})^\alpha \log(P_{xy}), & \text{if } GT(x,y) = 1 \\
(1 - GT(x,y))^\beta (P_{xy})^\alpha \log(1 - P_{xy}), & \text{otherwise}
\end{cases}
$}
\end{equation}

\subsubsection{Row-based segmentation prediction}
Row-based segmentation predicts a horizontal coordinate $\mathrm{E}\left(\hat{x}_{i}\right)$ for each row. For this task, our method uses the L1 loss function to optimize the predicted coordinates. Suppose $V$ denotes the effective vertical range of the lane line, $N_v$ is the number of effective rows, and $x_i$ is the ground truth horizontal coordinate. The loss function is defined as:
\begin{equation}
L_{\text{row}} = \frac{1}{N_v} \sum_{i \in V} \left| \mathrm{E}\left(\hat{x}_i\right) - x_i \right|
\end{equation}
\subsubsection{Vertical range prediction}
For the label of the lane line vertical range vector prediction, our method uses a \(1 \times Y\) vector representation. If the lane line passes through the $Y_i$ row, its value is 1; otherwise, it is 0. Thus, this can be regarded as a segmentation task. Our method optimizes it using the cross-entropy loss function. Suppose \(v_i\) denotes the predicted probability that the \(i\)-th row is positive, then the loss function is defined as:
\begin{equation}
L_{\text{range}} = \sum \left( -y_{gt}^i \log(v_i) - (1 - y_{gt}^i) \log(1 - v_i) \right)
\end{equation}

\subsubsection{Offset prediction}
Firstly, the generation of ground truth labels for offset prediction is required. For each lane line, the proposed method constructs the ground truth offset as the horizontal displacement relative to the lane centerline for each grid cell within a specified width. The predicted offsets are optimized using an L1 loss function. Let \(\delta_{xy}\) denote the ground truth offset at coordinate \((x,y)\), \(\hat{\delta}_{xy}\) the predicted offset, and \(\Omega\) the region surrounding the lane line. The loss function is then formulated as follows:

\begin{equation}
L_{\text{offset}} = \frac{1}{N_{\Omega}} \sum_{(x,y) \in \Omega} \left| \hat{\delta}_{xy} - \delta_{xy} \right|
\end{equation}

\subsubsection{Edge prediction}
For the task of edge enhancement, the method treats it as an edge segmentation problem, which is essentially a binary classification task. The labels at edge locations are assigned a value of 1, while non-edge locations are assigned 0. Accordingly, the cross-entropy loss function is employed for optimization. 

As discussed in Section \ref{sec:edge-enhancement}, the loss function for edge prediction is defined as:
\begin{equation}
L_{\text{edge}} = \sum_{i} \left( -e_{gt}^i \log(p_i) - (1 - e_{gt}^i) \log(1 - p_i) \right)
\end{equation}
where \( e_{gt}^i \) denotes the label of pixel \( i \), and \( p_i \) represents the predicted probability of the pixel being part of the edge.

\begin{figure}
    \centering
    \includegraphics[width=0.9\columnwidth]{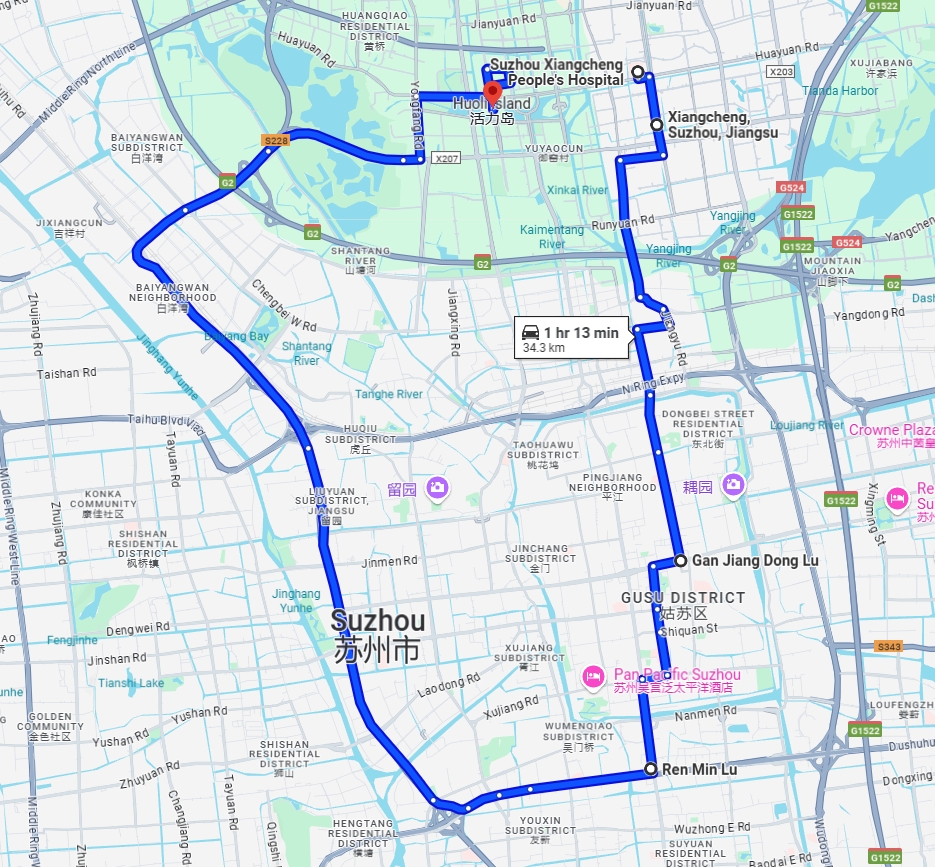}
    \caption{The route for dataset acquisition in Suzhou on the morning of October 26, 2023.\citep{googlemap}}
    \label{fig_suzhou}
\end{figure}

\begin{figure*}[htb]
    \centering
    \includegraphics[width=0.95\textwidth]{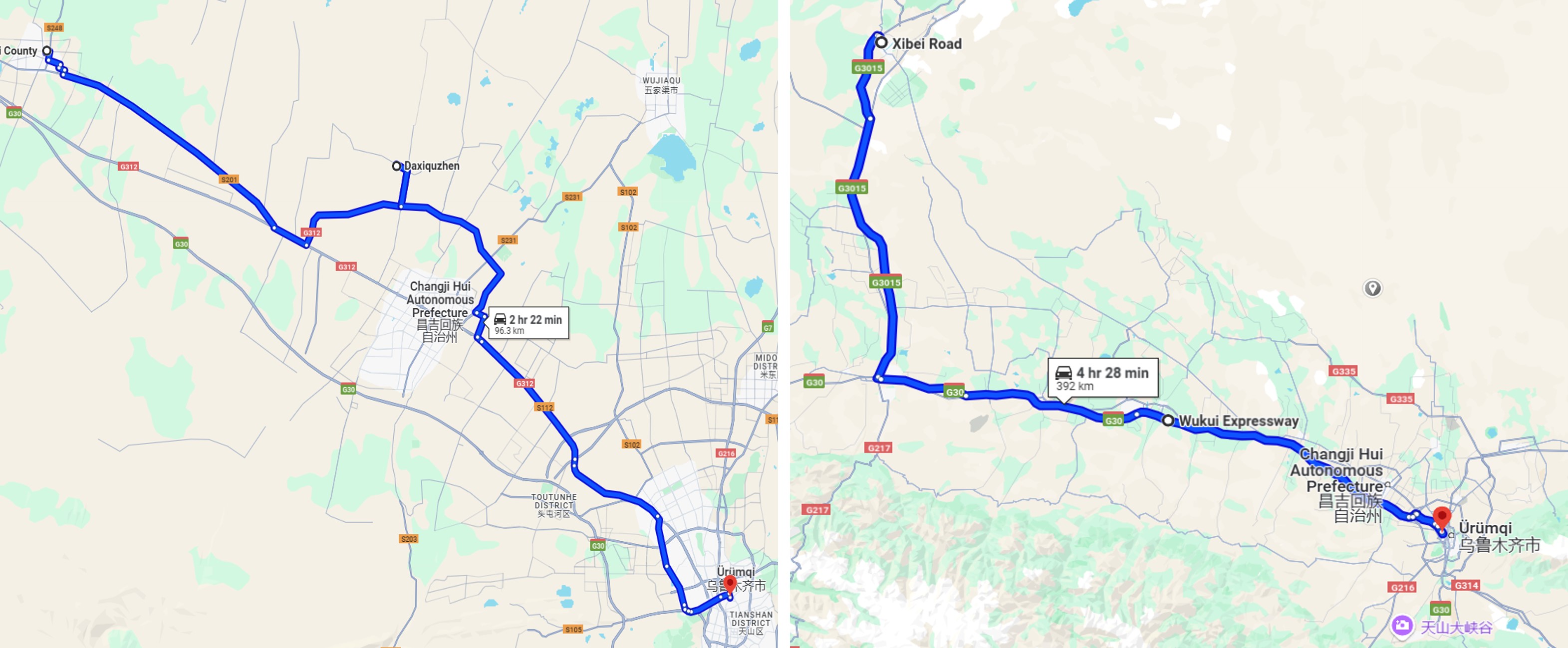}
    \caption{The routes for dataset acquisition in Urumqi, Xinjiang Autonomous Region: the left illustrates the morning of December 9, 2023, while the right depicts the evening of January 2, 2024.\citep{googlemap}}
    \label{fig_wlmq}
\end{figure*}

\subsubsection{Total loss}
To balance the influence of different tasks on the overall training of the network, a weighted fusion of the loss functions for each sub-task is employed. The total loss function is formulated as follows:

\begin{equation}
L_{\text{total}} = \alpha L_{hm} + \beta L_{\text{row}} + \gamma L_{\text{range}} + \delta L_{\text{offset}} + \varepsilon L_{\text{edge}}
\end{equation}
where the network achieves optimal performance when the weights are set as \(\alpha = 1\), \(\beta = 1\), \(\gamma = 1\), \(\delta = 0.4\), and \(\varepsilon = 0.1\).

\section{Experiments}
\subsection{Datasets}
Currently, there is no publicly available foggy weather lane detection dataset. To fill this gap, we created the FoggyLane dataset for lane detection in foggy weather. In addition, we generated two supplementary datasets, FoggyCULane and FoggyTusimple, by using fog modeling techniques on existing CULane \citep{pan2018spatial} and Tusimple \citep{Tusimple} datasets to test the generalizability of our methods.

\begin{figure*}[htb]
    \centering
    \includegraphics[width=0.6\textwidth]{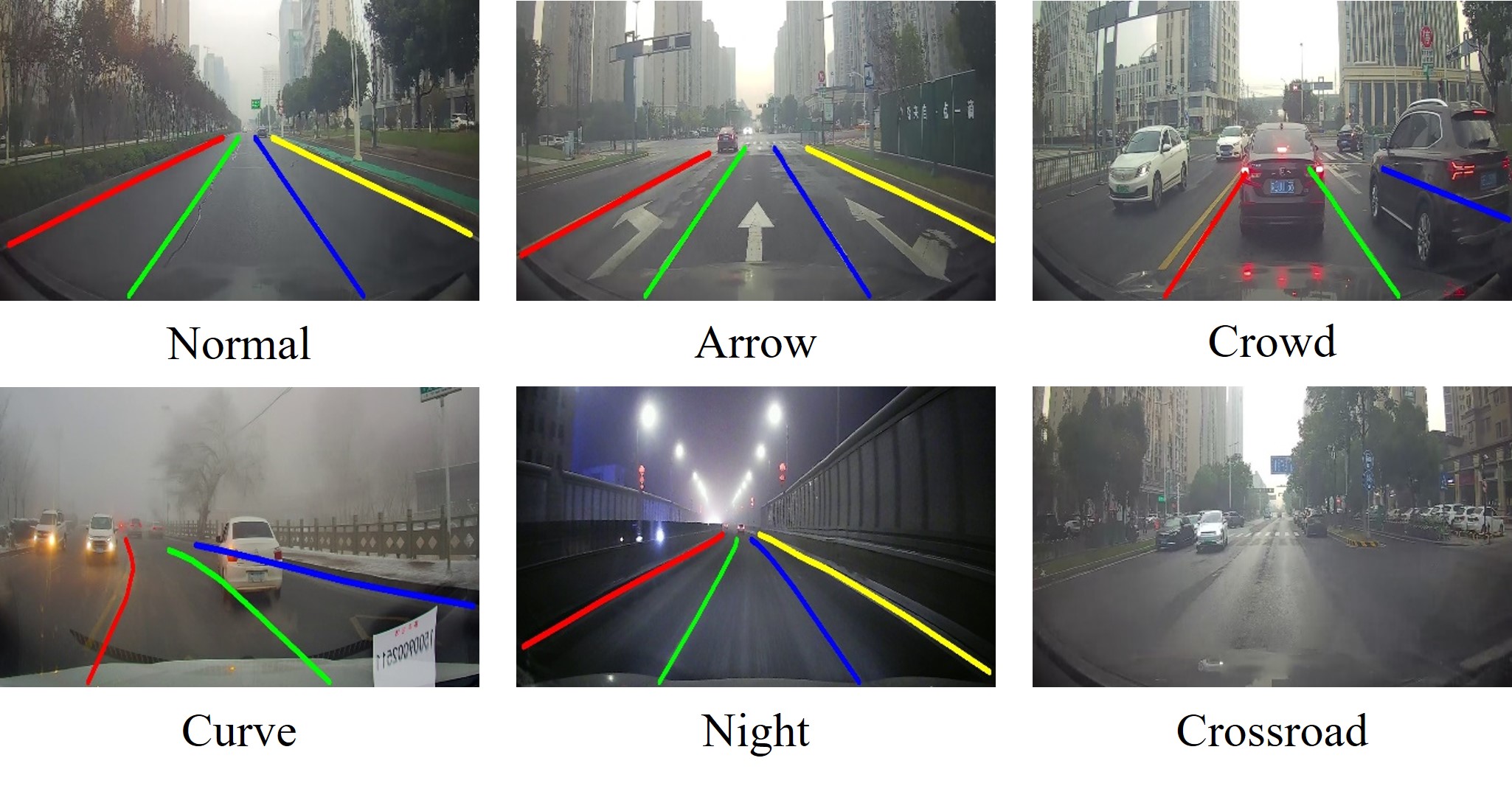}
    \caption{The different road types of FoggyLane images examples.}
    \label{fig_dataset}
\end{figure*}

\begin{table*}
    \caption{Comparison Between Our Dataset and Existing Lane Detection Datasets\label{tab:dataset comparison}}
    \begin{adjustbox}{width=\textwidth}
        \begin{tabular}{cccccc}
\bottomrule
Dataset&Frames&Multi-traffic scenarios&Instance-level annatation&Number of foggy datasets&Resolution\\
\midrule
Caltech-Lanes \citep{Caltech-Lanes} & 1.2k & \textcolor{red}{\ding{55}} & \textcolor{green}{\ding{51}} & None & 640 x 480 \\
Tusimple \citep{Tusimple} & 6.4k & \textcolor{red}{\ding{55}} & \textcolor{green}{\ding{51}} & None & 1280 x 720 \\
LLAMAS \citep{LLAMAS} & 100k & \textcolor{green}{\ding{51}} & \textcolor{green}{\ding{51}} & None & 1276 x 717 \\
Appolloscape \citep{Apolloscape} & 110k & \textcolor{green}{\ding{51}} & \textcolor{red}{\ding{55}} & None &3384 x 2710 \\
BDD100k \citep{yu2020bdd100k} & 100k & \textcolor{green}{\ding{51}} & \textcolor{red}{\ding{55}} & None & 1280 x 720 \\
CULane \citep{pan2018spatial} & 133k & \textcolor{green}{\ding{51}} & \textcolor{green}{\ding{51}} & None & 1640 x 590 \\
CurveLanes \citep{xu2020curvelane} & 150k & \textcolor{green}{\ding{51}} & \textcolor{green}{\ding{51}} & None & 1280 x 720 \\
FoggyLane (ours) & 1.423k & \textcolor{green}{\ding{51}} & \textcolor{green}{\ding{51}} & \textcolor{blue}{1.423k} & 1640 x 590 \\
\bottomrule
        \end{tabular}
    \end{adjustbox}
\end{table*}

\subsubsection{FoggyLane}
Existing lane detection algorithms for intelligent connected vehicle ADAS systems perform well under favorable weather conditions, but their performance significantly deteriorates in adverse conditions like fog, which reduces visibility and increases the likelihood of traffic accidents. Current benchmark datasets, including CULane\citep{pan2018spatial}, Tusimple\citep{Tusimple}, and LLAMAS\citep{LLAMAS}, lack foggy weather data. To address this, we developed FoggyLane, a dataset specifically designed for foggy conditions. Data was collected using a monocular forward-facing camera at various times and locations, including Suzhou and Urumqi, China, between October 2023 and January 2024. The dataset includes 1,086 images at various resolutions, which were resized to $1640 \times 590$ for consistency.

Since foggy conditions are rare, we extended the dataset by incorporating foggy driving scenes from two YouTube videos\citep{YouTube}, resulting in an additional 337 images. The final FoggyLane dataset contains 1,423 images, covering six road scenes: Normal, Arrow, Crowd, Curve, Night, and Crossroad. The dataset was partitioned into training, validation and testing sets, with stratification to maintain a consistent distribution of scene types across both subsets.

Comparing FoggyLane with existing datasets, such as Caltech-Lanes\citep{Caltech-Lanes} and Tusimple\citep{Tusimple}, which focus on single traffic scenarios, and LLAMAS\citep{LLAMAS}, which uses high-precision map annotations, we found that these datasets do not cover foggy weather scenarios. While ApolloScape \citep{Apolloscape} and BDD100k \citep{yu2020bdd100k} include multiple traffic scenarios, they lack instance-level annotations. Similarly, although CULane \citep{pan2018spatial} and CurveLanes \citep{xu2020curvelane} cover a wide range of traffic scenes, they do not include foggy conditions. Models trained on datasets limited to clear weather struggle in fog, where lane boundaries are blurred and image quality reduced. To fill this gap, FoggyLane offers a thorough solution for lane detection in foggy conditions with instance-level annotations.

Our dataset has two key advantages:
\begin{itemize}
\item [1)] Pioneering foggy lane detection dataset: FoggyLane is a dataset specifically designed to address foggy weather scenarios. It covers a range of challenging conditions, including foggy congested roads, foggy curves, and foggy night roads. This dataset is crucial for testing the robustness of ADAS under adverse conditions.
\item [2)] Strict annotation standards: The dataset follows strict annotation guidelines, including marking the center of lanes, connecting lane markings at intersections, and prioritizing solid lines over grid lines, etc.\citep{zhang2024robust}. These rigorous annotations can significantly improve the accuracy and consistency of lane detection in foggy weather, thus enhancing the stability and safety of autonomous driving systems and providing a solid data foundation for the reliable deployment of autonomous technology.
\end{itemize}

\begin{figure*}[htb]
    \centering
    \includegraphics[width=\textwidth]{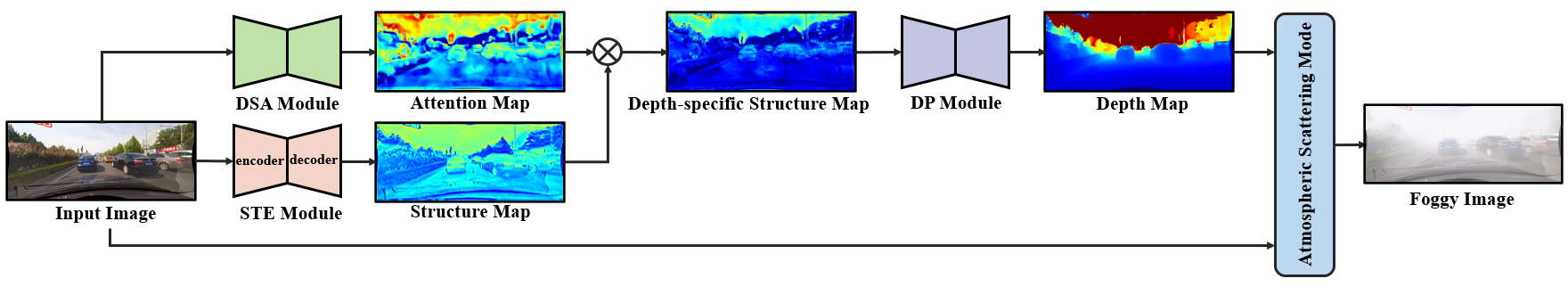}
    \caption{Networks of foggy image generation. $\otimes$ represents element-wise multiplication.}
    \label{S2R}
\end{figure*}

\begin{figure*}[htb]
    \centering
    \includegraphics[width=\textwidth]{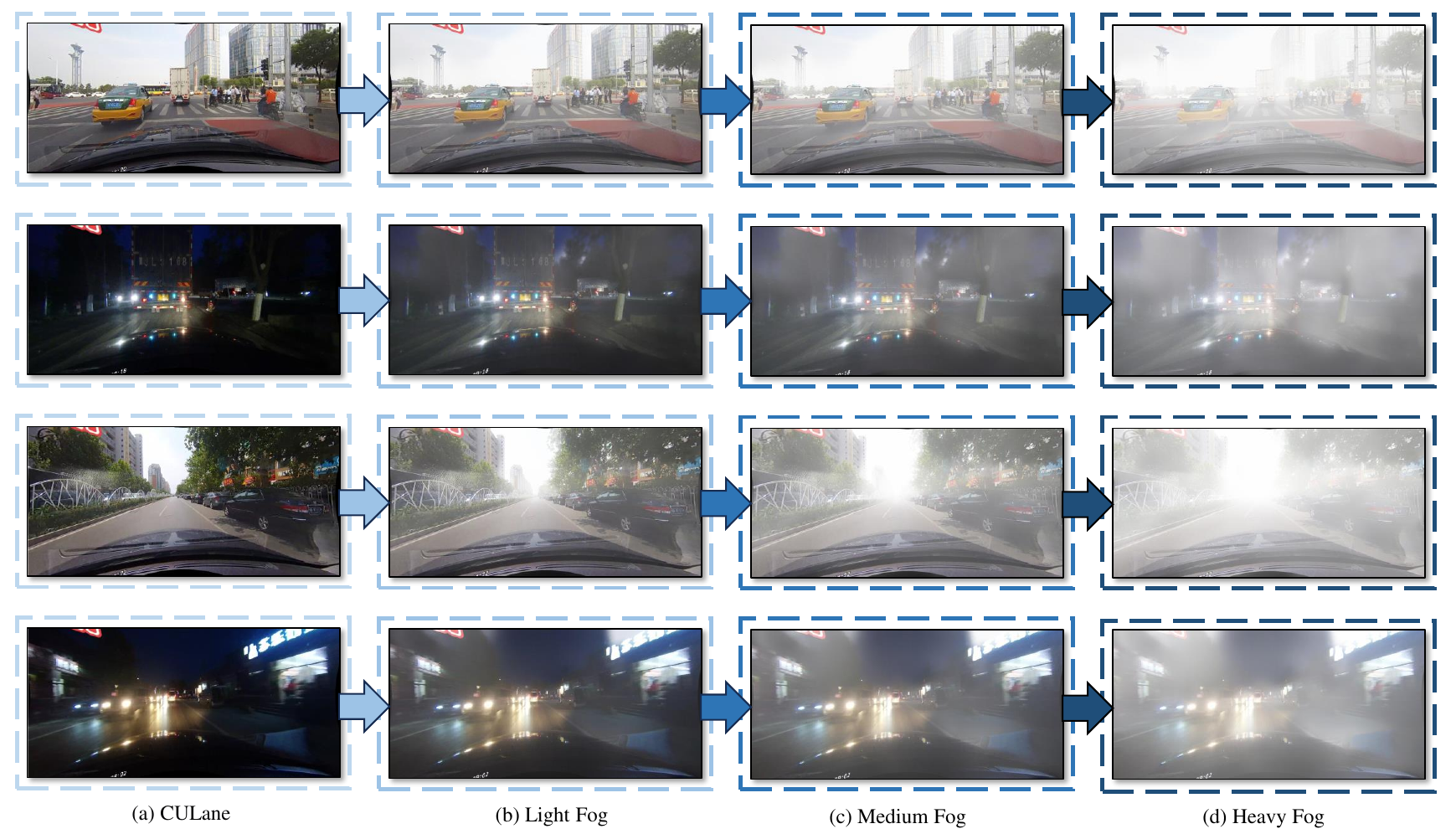}
    \caption{The FoggyCULane dataset examples. (a) is the original CULane dataset. And (b) to (d) are the foggy images of different concentrations with $\beta$ values of 2, 4, and 8, respectively.}
    \label{culane}
\end{figure*}

\subsubsection{FoggyCULane and FoggyTusimple}
Currently, numerous datasets are available for lane detection. Among them, we select two widely adopted and representative benchmarks: CULane\citep{pan2018spatial} and TuSimple\citep{Tusimple}. The CULane dataset \citep{pan2018spatial} serves as an extensive and demanding benchmark, encompassing various scenarios such as highways, urban roads, and rural regions in Beijing. In contrast, the TuSimple dataset \citep{Tusimple} primarily comprises daytime driving videos collected on American highways, featuring a variety of lighting conditions and complex traffic situations. Based on these datasets, we generate synthetic foggy versions, named FoggyCULane and FoggyTuSimple, to facilitate evaluation under adverse weather conditions.

The method for generating foggy images is inspired by the model \citep{nie2022sensor}, and the complete generation process is shown in Fig. \ref{S2R}. To be specific, the atmospheric scattering model can effectively utilize prior information to analyze the degradation mechanisms of image quality in foggy conditions. It models foggy images as a combination of scattered light and direct light based on optical principles, ultimately synthesizing a clear image into a foggy one. The mathematical modeling equation defined by Koschmieder \citep{harald1924theorie} is represented as:
\begin{equation}
I(x)=J(x) t(x)+A(1-t(x)) 
\end{equation}
where $x$ indicates a certain pixel of the image, $J(x)$ refers to the original clear image of the object and $I(x)$ refers to the foggy image. $A$ is the atmospheric light value at infinity and $t(x) $ can be further defined as:
\begin{equation}
t(x)=e^{-\beta d(x)}
\label{eq:beta and d}
\end{equation}
where $\beta$ is the extinction coefficient, which is closely related to the concentration, size, type and distribution of particulate matter. $d(x)$ represents the distance between the object and the viewer. As this distance increases, the transmittance decreases, resulting in thicker fog.

To calculate the corresponding $d(x)$, we use the S2R-DepthNet\citep{chen2021S2R} that has good generalization performance across various applications. The S2R-DepthNet consists of three modules that decouple the structural and textural information of the input image. Then they suppress information irrelevant to depth and use structural information for depth prediction. The Structure Extraction Module (STE) breaks down the image to capture domain-independent structural details. The Depth-specific Attention (DSA) generates an attention map from the original image. Once processed through the DSA and STE Modules, as depicted in Fig. \ref{S2R}, the attention map and structural map are multiplied to produce a refined and all-encompassing depth-specific structure map. The Depth Prediction (DP) module is then responsible for estimating the depth.

According to the transmittance formula Eq. (\ref{eq:beta and d}), we use the dark channel prior\citep{he2011darkchannel} to calculate the average atmospheric light value $A$. In the non-sky area of an outdoor clear image, there is usually at least one very small or close to zero value in the RGB image for each pixel, which is called the dark channel of that pixel. Therefore, for any image $I$, the mathematical formula for its corresponding dark channel image $I_{\text {dark}}(x) $ can be defined as:
\begin{equation}
I_{\text {dark }}(x)=\min _{c \in R, G, B}\left(\min _{y \in \Omega(x)} I_c(y)\right)
\end{equation}
where $I_{\text{c}}(y)$ is the value of the c-th channel of image $I$ at pixel point $y$, and $\omega(x)$ is a small region around pixel point $x$. In the dark channel, find the top 0.1\% brightest pixels and take the maximum value of their RGB channels as an approximation of the average atmospheric light value $A$. After obtaining the transmittance and the average atmospheric light value, the images of light fog, medium fog, and heavy fog can be obtained by fine-tuning the extinction coefficient $\beta$. The original images and the foggy images generated under different $\beta$ values are displayed in Figs. \ref{culane}.
\begin{table*}[htb]
\caption{Comparison with Advanced Lane Detection on FoggyLane dataset.}
\label{table_foggylane_result}
\begin{adjustbox}{width=\textwidth}
\begin{tabular}{lclccccccc>{\columncolor[HTML]{F2F2F2}}c>{\columncolor[HTML]{F2F2F2}}c}
\Xhline{1pt}
\textbf{Method} & \textbf{Venue} & \textbf{Backbone} & \textbf{F1@50} & \textbf{F1@65}& \textbf{F1@75} & \textbf{F1@85}& \textbf{mF1}&\textbf{Gflops(G)} & \textbf{FPS}\\
\hline
%\multicolumn{12}{l}{\textbf{Segmentation-based}} \\
SCNN\citep{pan2018spatial}  & \textit{AAAI 2018} & VGG16   & 84.01 & 62.08& 56.34 &12.76 &41.78& 328.4 & 18.7\\
UFLD\citep{qin2020ultra} & \textit{ECCV 2020}& ResNet18  & 81.76&56.54 & 32.32&6.50& 38.24& 8.4 & 327.3\\
UFLD\citep{qin2020ultra} & \textit{ECCV 2020} & ResNet34  & 82.75&61.25 & 34.44 &7.89&40.23& 16.9 & 176.9\\
LaneATT\citep{tabelini2021keep} & \textit{CVPR 2020}& ResNet18   & 81.98 &59.61& 34.78&7.69&40.06 & 9.3 & 247.7\\
LaneATT\citep{tabelini2021keep} & \textit{CVPR 2020}& ResNet34   & 84.32 &63.08& 38.91 &10.76&42.63& 18.0 & 167.1 \\
%\multicolumn{12}{l}{\textbf{Curve-based}} \\
LSTR\citep{liu2021end} & \textit{WACV 2021} & ResNet34   & 83.79 &68.23& 58.29&25.32& 48.91& - & 124.4\\
RESA\citep{zheng2021resa}  & \textit{AAAI 2021}& ResNet34  & 85.19 &65.89& 61.89 &14.57&48.75& 41.0 & 76.7 \\
RESA\citep{zheng2021resa}  & \textit{AAAI 2021} & ResNet50  & 86.24 &69.22 & 65.27 &16.55&52.34& 43.0 & 53.0\\
CondLane\citep{liu2021condlanenet}  & \textit{ICCV 2021} & ResNet18   & 90.38 &75.08& 47.50 &47.50&10.96& 10.2 & 194.7 \\
CondLane\citep{liu2021condlanenet}  & \textit{ICCV 2021} & ResNet34  & 90.58 &78.45& 56.48 &15.59&52.25& 19.6 & 136.5 \\
CLRNet\citep{zheng2022clrnet}  & \textit{CVPR 2022}& ResNet34  & 91.56 &82.43& 65.61 &30.20&58.14 & 21.5 & 90.0\\
CLRNet\citep{zheng2022clrnet}  & \textit{CVPR 2022} & DLA34  & 92.09 &83.98& 72.15&41.61&62.28 & 18.5 & 81.3 \\
FLAMNet\citep{ran2023flamnet}  & \textit{IEEE TITS 2023}& ResNet34 & 89.94 &78.90& 55.32&12.96&50.99 & 30.1 & 32.0 \\
FLAMNet\citep{ran2023flamnet}  & \textit{IEEE TITS 2023}& DLA34   & 91.16 &82.77& 64.95&24.22& 56.70& 21.7 & 31.7\\
DBNet\citep{dai2024dbnet} & \textit{IEEE TIV 2024} & ResNet18  & 89.61 &81.27& 65.27 &26.15&56.52& 22.6 & 147.6\\
DBNet\citep{dai2024dbnet}  & \textit{IEEE TIV 2024} & ResNet34  & 90.02 &83.44& 71.66 &40.51&60.94& 32.1 & 129.0\\
HWLane\citep{tits3}  & \textit{IEEE TITS 2024}& Res34   & 91.99 &84.23& 68.08&35.27& 34.38& 18.6 & 70.0\\
PolarRCNN\citep{tits4}  & \textit{IEEE TITS 2025} & ResNet34  & 91.07 &79.38& 57.44 &14.74&52.42& 19.0 & 171.6\\
PolarRCNN\citep{tits4}  & \textit{IEEE TITS 2025} & DLA34  & 90.02 &83.44& 71.66 &40.51&51.97& 16.0 & 150.4\\
\hline
PDTNet(Ours) & & SwinGFFM-t  &92.34&84.65& 70.99 &35.77&60.82& 12.0 & 70.4 / 304.7*\\
PDTNet(Ours) & \multirow{-2}{*}{-}& SwinGFFM-s   & \textbf{\textcolor{blue}{95.04(2.95$\uparrow$)}} &\textbf{\textcolor{blue}{88.61(4.63$\uparrow$)}}& \textbf{\textcolor{blue}{77.32(5.17$\uparrow$)}} &\textbf{\textcolor{blue}{45.02(3.41$\uparrow$)}}&\textbf{\textcolor{blue}{65.60(3.32$\uparrow$)}}& 20.3 & 38.2 / 192.9*\\
\Xhline{1pt}
\end{tabular}
\end{adjustbox}
\begin{tablenotes}
    \footnotesize
    \item Note: For a fairer comparison, we re-evaluated the FPS of the source code available detectors using one NVIDIA GeForce RTX 3090 GPU on the same machine. * indicates the FPS after acceleration with TensorRT.
\end{tablenotes}
\end{table*}

\subsection{Evaluation Metric}
Lane detection is typically approached as a semantic segmentation problem, assessed using common metrics. However, due to its distinct characteristics, lane detection also requires specialized evaluation metrics. Adhering to SCNN’s \citep{pan2018spatial} framework, we link labeled or predicted lane points to form line segments and extend them into 30-pixel-wide regions. Subsequently, we compute the Intersection over Union (IoU), and if the IoU value exceeds a predefined threshold, typically set at 0.5, the prediction is considered correct. From this, we calculate Precision, Recall, and F1-score using the following formulas:
{\begin{align}
F1 = \frac{2\times Precision \times Recall}{Precision + Recall}
\end{align}}
{\begin{align}
Precision = \frac{TP}{TP + FP} 
\end{align}}
{\begin{align}
Recall = \frac{TP}{TP + FN} 
\end{align}}

\begin{figure*}[htb]
    \centering
    \hspace*{-1cm} % 向左移动的距离
    \begin{minipage}{\textwidth}
        \includegraphics[width=0.95\textwidth]{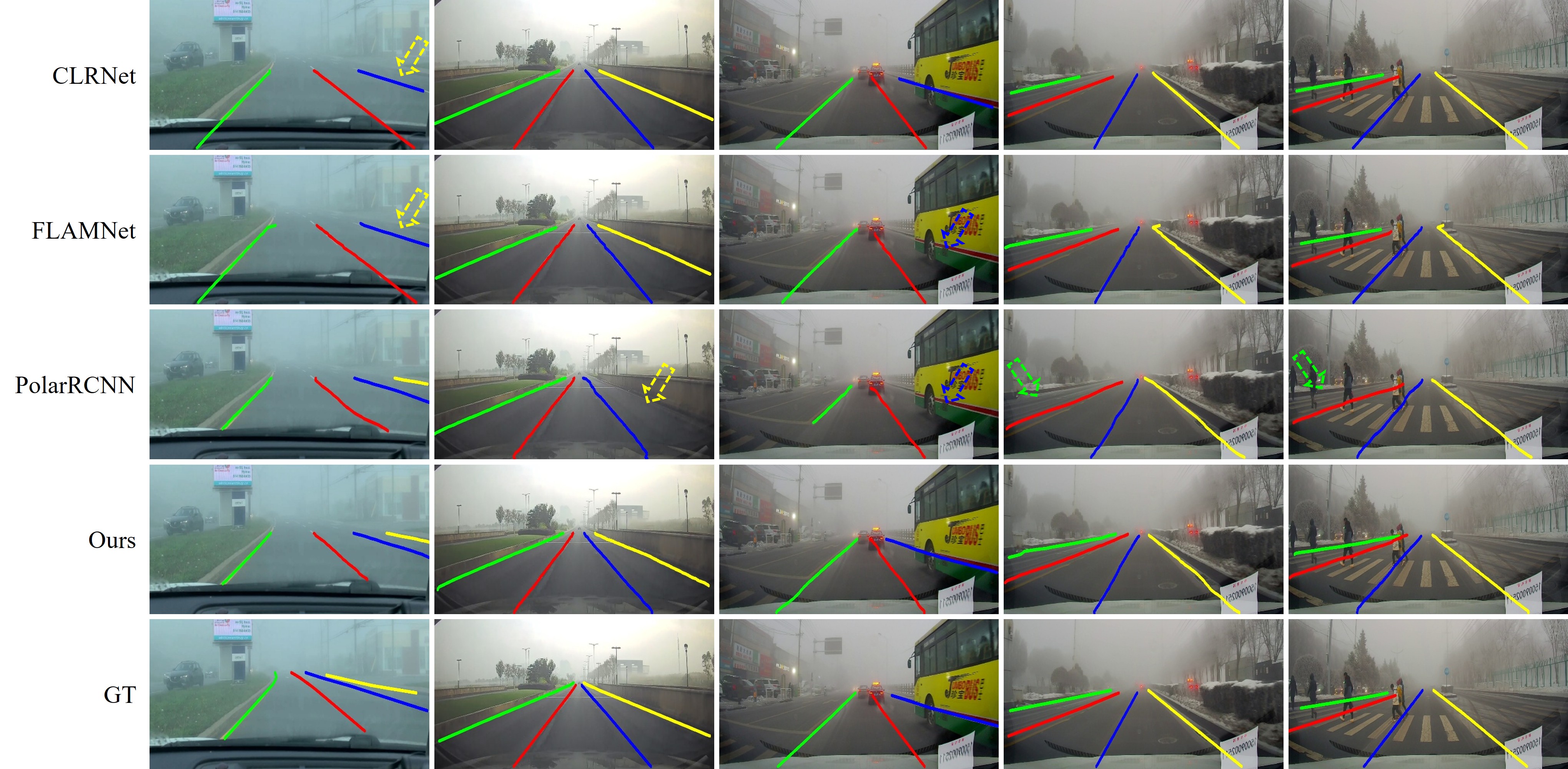}
    \end{minipage}
    \caption{Visualization results of CLRNet\citep{zheng2022clrnet}, FLAMNet\citep{ran2023flamnet}, PolarRCNN\citep{tits4} and our method on FoggyLane testing set. The arrows indicate instances of lane lines that were not correctly detected by the corresponding method.}
    \label{fig_foggylane_results}
\end{figure*}

\begin{table*}[htb]
\caption{Comparison with Advanced Lane Detection on FoggyCULane dataset}
\label{table_FoggyCULane_result}
\begin{adjustbox}{width=\textwidth}
\begin{tabular}{lclcccccccccc}
\Xhline{1pt}
\textbf{Method} & \textbf{Venue} & \textbf{Backbone} & \textbf{Normal} & \textbf{Crowded} & \textbf{Dazzle light} & \textbf{Shadow} & \textbf{No line} & \textbf{Arrow} & \textbf{Curve} & \textbf{Cross} & \textbf{Night} & \textbf{Total} \\
\hline
%\multicolumn{2}{l}{\textbf{Segmentation-based}} \\
SCNN\citep{pan2018spatial}           & \textit{AAAI 2018}  & VGG16   & 91.07 & 71.32 & 67.23 & 65.23 & 49.14 & 88.06 & 67.07 & 1344 & 70.78 & 72.94 \\
%\hline
UFLD\citep{qin2020ultra}           & \textit{ECCV 2020} & ResNet18  & 87.57 & 65.47 & 55.21 & 60.38 & 38.66 & 82.26 & 57.29 & 1849 & 60.61 & 67.69 \\
UFLD\citep{qin2020ultra}           & \textit{ECCV 2020} & ResNet34 & 89.10 & 68.29 & 60.91 & 62.03 & 41.98 & 85.17 & 63.88 & 2371 & 70.13 & 70.13 \\
%\hline
LaneATT\citep{tabelini2021keep}        &\textit{CVPR 2020} & ResNet18  & 90.52 & 71.52 & 65.70 & 68.92 & 47.31 & 86.32 & 62.34 & 870& 67.64 & 74.30 \\
LaneATT\citep{tabelini2021keep}        &\textit{CVPR 2020} & ResNet34  & 91.68 & 72.93 & 67.23 & 65.23 & 49.14 & 88.06 & 67.07 & 1344 & 70.78 & 75.53 \\
%\hline
LSTR\citep{liu2021end}           & \textit{WACV 2021}  & ResNet34  & 85.79 & 63.45 & 56.73 & 58.72 & 38.20 & 79.31 & 55.34 & 1068 & 57.62 & 65.64 \\
%\hline
RESA\citep{zheng2021resa}           & \textit{AAAI 2021} & ResNet34  & 90.97 & 72.32 & 67.31 & 74.01 & 46.52 & 86.32 & 65.26 & 1536 & 73.62 & 74.97 \\
RESA\citep{zheng2021resa}          & \textit{AAAI 2021} & ResNet50 & 92.09 & 73.49 & 67.40 & 73.87 & 47.92 & 87.48 & 69.28 & 1722 & 69.85 & 75.19 \\
%\hline
CondLane\citep{liu2021condlanenet}    & \textit{ICCV 2021}  & ResNet18  & 91.56 & 74.41 & 71.12 & 77.89 & 48.59 & 86.73 & 71.34 & 1262 & 71.00 & 76.46 \\
CondLane\citep{liu2021condlanenet}    & \textit{ICCV 2021}  & ResNet34  & 91.76 & 75.55 & 69.62 & 74.62 & 50.01 & 87.40 & 71.29 & 1427 & 72.11 & 77.06 \\
%\hline
CLRNet\citep{zheng2022clrnet}         & \textit{CVPR 2022} & ResNet34  & 92.96 & 76.32 & 73.40 & 79.06 & 51.46 & 89.39 & 71.83 & 1301 & 73.64 & 78.36 \\
CLRNet\citep{zheng2022clrnet}         & \textit{CVPR 2022} & DLA34     & 93.23 & 77.64 & 73.90 & 79.23 & 52.11 & 89.45 & 72.48 & 1109 & 73.63 & 79.04 \\
%\hline
FLAMNet\citep{ran2023flamnet}        & \textit{IEEE TITS 2023} & ResNet34  & 92.60 & 76.61 & 71.42 & 79.81 & 50.89 & 89.04 & 71.47 & 1073 & 74.17 & 78.45 \\
FLAMNet\citep{ran2023flamnet}        & \textit{IEEE TITS 2023} & DLA34     & 92.98 & 77.26 & 71.84 & 79.76 & 51.29 & 89.34 & 71.38 & 1143 & 73.93 & 78.77 \\
%\hline
%\multicolumn{12}{l}{\textbf{Curve-based}} \\
DBNet\citep{dai2024dbnet}          & \textit{IEEE TIV 2024} & ResNet18  & 84.91 & 69.27 & 60.86 & 67.94 & 45.69 & 79.04 & 67.50 & 1045 & 68.57 & 71.33 \\
DBNet\citep{dai2024dbnet}          & \textit{IEEE TIV 2024} & ResNet34 & 85.61 & 69.35 & 61.22 & 66.50 & 45.73 & 79.19 & 66.51 & \textbf{\textcolor{blue}{837}} & 69.93 & 71.87 \\
HWLane\citep{tits3}  & \textit{IEEE TITS 2024}& Res34   & 90.85 &73.32& 67.18&75.32& 47.18& 87.14 & 63.38&1456&69.20&74.61\\
PolarRCNN\citep{tits4}  & \textit{IEEE TITS 2025} & ResNet34  & 93.34 &77.76& \textbf{\textcolor{blue}{75.26}}&79.76& 53.25& 89.62 & \textbf{\textcolor{blue}{75.94}}&1289&74.54&78.68\\
PolarRCNN\citep{tits4}  & \textit{IEEE TITS 2025} & DLA34  & 92.90 &77.50& 74.03&80.89& 51.71& 89.76 & 73.02&1226&74.21&79.13\\
\hline

PDTNet(Ours)           &                     & SwinGFFM-t    & 92.68 & 77.50 & 71.52 & 80.42 & 50.95 & 88.92 & 72.08 & 950  & 73.11 & 78.66 \\
PDTNet(Ours)         &  \multirow{-2}{*}{-}& SwinGFFM-s    & \textbf{\textcolor{blue}{93.40}} & \textbf{\textcolor{blue}{78.63(0.87$\uparrow$)}} & 73.89 & \textbf{\textcolor{blue}{82.74(1.85$\uparrow$)}} & \textbf{\textcolor{blue}{53.75(0.50$\uparrow$)}} & \textbf{\textcolor{blue}{89.82(0.37$\uparrow$)}} & 74.09 & 1146 & \textbf{\textcolor{blue}{74.97(0.76$\uparrow$)}} & \textbf{\textcolor{blue}{79.85(0.72$\uparrow$)}} \\
\Xhline{1pt}
\end{tabular}
\end{adjustbox}
\begin{tablenotes}
    \footnotesize
    \item Note: Our method attains state-of-the-art (SOTA) results in multiple scenarios including Normal, Crowded, Shadow, No-line, Arrow and Night conditions, as well as in the overall (Total) performance. 
\end{tablenotes}
\end{table*}

\begin{figure*}[htb]
    \centering
    \hspace*{-1cm} % 向左移动的距离
    \begin{minipage}{0.95\textwidth}
        \includegraphics[width=\textwidth]{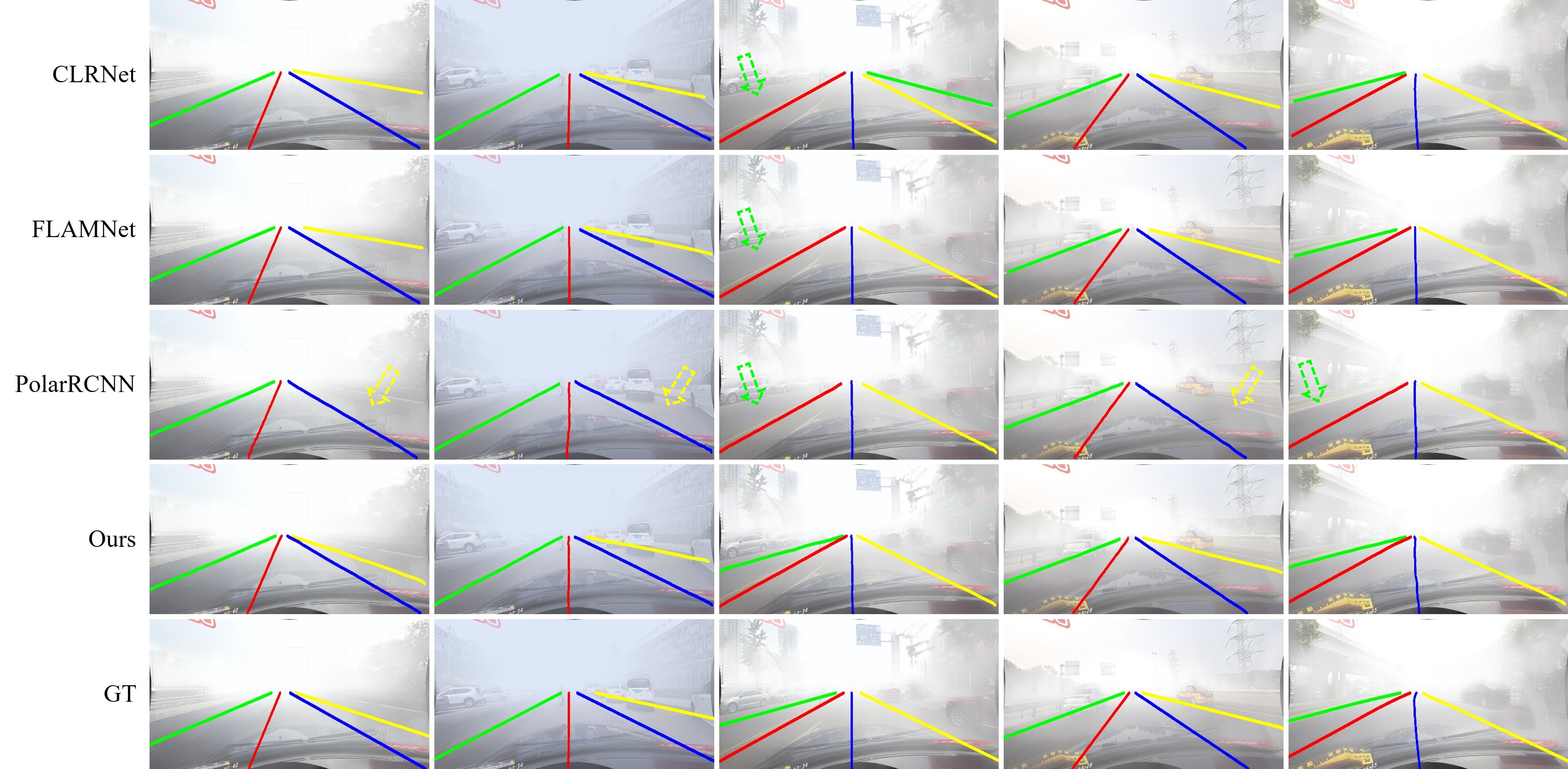}
    \end{minipage}
    \caption{Visualization results of CLRNet\citep{zheng2022clrnet}, FLAMNet\citep{ran2023flamnet}, PolarRCNN\citep{tits4} and our method on FoggyCULane testing set. The arrows indicate instances of lane lines that were not correctly detected by the corresponding method.}
    \label{fig_foggyculane_results}
\end{figure*}
For the FoggyTusimple dataset, we use the same evaluation metrics as the Tusimple dataset. The official Tusimple dataset includes three standard metrics: false positive rate (FPR), false negative rate (FNR), and accuracy. The formula for calculating accuracy is as follows:

\begin{align}
\textit{Accuracy} = \frac{\sum_{\textit{clip}} C_{\textit{clip}}}{\sum_{\textit{clip}} S_{\textit{clip}}}
\end{align}
where $C_{\textit{clip}}$ represents the number of correctly predicted lane points in the image, while $S_{\textit{clip}}$ denotes the total number of ground truth lane points.

\begin{table*}[t]
\caption{Comparison with Advanced Lane Detection on FoggyTusimple dataset.}
\centering
\label{table_FoggyTusimple_result}
\begin{adjustbox}{width=0.85\textwidth}
\begin{tabular}{lcccccc}
\Xhline{1pt}
\multicolumn{1}{l}{\textbf{Method}} & \textbf{Venue }
& \textbf{Backbone}      & \textbf{F1(\%)} & \textbf{ACC(\%)}  & \textbf{FP(\%)} & \textbf{FN(\%)}    \\
\hline
SCNN\citep{pan2018spatial}                        & \textit{AAAI 2018}                   & VGG16     & 94.64 & 95.89 & 5.87 & 4.78 \\
UFLD\citep{qin2020ultra}             & \textit{ECCV 2020} & ResNet18  & 94.57 & 93.60 & 3.49 & 7.59 \\ 
UFLD\citep{qin2020ultra}             &  \textit{ECCV 2020} & ResNet34  & 95.05 & 95.10 & 4.84 & 5.08 \\ 
LaneATT\citep{tabelini2021keep}         & \textit{CVPR 2020} & ResNet18  & 95.77 & 95.78 & 3.83 & 3.53 \\ 
LaneATT\citep{tabelini2021keep}         & \textit{CVPR 2020} & ResNet34  & 96.15 & 95.50 & 4.59 & 3.06 \\ 
LSTR\citep{liu2021end}             & \textit{WACV 2021} & ResNet34  & 94.70 & 95.58 & 5.74 & 4.81 \\ 
RESA\citep{zheng2021resa}            & \textit{AAAI 2021} & ResNet34  & 95.22 & 95.60 & 4.63 & 4.95 \\
RESA\citep{zheng2021resa}            & \textit{AAAI 2021} & ResNet50  & 95.51 & 95.83 & 4.53 & 4.43 \\
CondLane\citep{liu2021condlanenet}        & \textit{ICCV 2021} & ResNet18  & 95.41 & 94.81 & 3.80 & 5.46 \\ 
CondLane\citep{liu2021condlanenet}        &  \textit{ICCV 2021}& ResNet34  & 95.65 & 94.69 & 3.63 & 5.13 \\ 
CLRNet\citep{zheng2022clrnet}           & \textit{CVPR 2022} & ResNet18  & 96.48 & 95.64 & 3.89 & 3.11 \\
CLRNet\citep{zheng2022clrnet}           & \textit{CVPR 2022} & ResNet34  & 96.23 & 95.94 & 4.34 & 3.17 \\
FLAMNet\citep{ran2023flamnet}         & \textit{IEEE TITS 2023} & ResNet18  & 96.52 & 95.65 & 3.84 & 3.09 \\
FLAMNet\citep{ran2023flamnet}         & \textit{IEEE TITS 2023} & ResNet101 & 96.34 & 95.98 & 4.45 & \textbf{\textcolor{blue}{2.82}}\\ 
DBNet\citep{dai2024dbnet}           & \textit{IEEE TIV 2024} & ResNet18  & 95.71 & 93.89 & 3.53 & 5.12 \\ 
DBNet\citep{dai2024dbnet}           & \textit{IEEE TIV 2024} & ResNet34  & 95.61 & 93.72 & 3.48 & 5.38 \\ 
HWLane\citep{tits3}  & \textit{IEEE TITS 2024}& Res34   & 95.54 &95.91& 4.32&4.60\\
PolarRCNN\citep{tits4}  & \textit{IEEE TITS 2025} & ResNet34  & 96.33 &95.91& 4.50 &2.77\\
PolarRCNN\citep{tits4}  & \textit{IEEE TITS 2025} & DLA34  & 96.37 &94.99& 3.21 &4.09\\
\hline
PDTNet(Ours)            &                           & SwinGFFM-t & 96.63 & 95.86 & 2.77 & 4.04 \\ 
PDTNet(Ours)            &  \multirow{-2}{*}{-}      & SwinGFFM-s & \textbf{\textcolor{blue}{96.95(0.43$\uparrow$)}}& \textbf{\textcolor{blue}{96.10(0.12$\uparrow$)}} & \textbf{\textcolor{blue}{2.32(0.89$\downarrow$)}} & 3.83 \\ \Xhline{1pt}
\end{tabular}
\end{adjustbox}
\end{table*}

\subsection{Implementation Details}
We conducted experiments on the NVIDIA GeForce RTX 3090 platform. The SwinGFFM backbone network was pre-trained on the COCO dataset for 500 epochs. During lane detection training, we loaded the pre-trained weights and resized input images to 800 × 320. Various data augmentation techniques, including rotation, flipping, brightness adjustment, and random scaling, were applied. For full network parameters and training details, please refer to our open-source code.

To benchmark against existing methods\citep{pan2018spatial,qin2020ultra,liu2021condlanenet,zheng2021resa,tabelini2021keep,liu2021end,zheng2022clrnet,ran2023flamnet,dai2024dbnet,tits4}, we followed the training settings provided in the respective papers. For FoggyCULane and FoggyTusimple, we adopted the training parameters from the original CULane and Tusimple datasets, while for FoggyLane, we extended the training epochs to ensure reliable results

\subsection{Quantitative Results}

\subsubsection{Results on FoggyLane}We evaluated our proposed method on the FoggyLane dataset and compared it with other leading lane detection methods. As shown in the Tabel \ref{table_foggylane_result}, we mainly compared the methods in terms of F1 scores, computational complexity (GFlops), processing speed (FPS). 

From the Table \ref{table_foggylane_result}, it is evident that our method achieves state-of-the-art performance on the foggy lane detection dataset FoggyLane, with F1@50, F1@65, F1@75, and F1@85 scores of 95.04, 88.61, 77.32, and 45.02, respectively. These results surpass the best-performing alternative methods by margins of 2.95\%, 4.63\%, 5.17\%, and 3.41\%. The mean F1 (mF1) score demonstrates a 3.32\% improvement over the strongest competing approach. Notably, the SwinGFFM-t variant achieves an F1@50 score of 92.34, which even exceeds the performance of DLA34-based CLRNet, previously the top-performing method among comparative approaches.

\subsubsection{Results on FoggyCULane}We compared our method with other advanced lane detection methods on the FoggyCULane dataset. As shown in the Table \ref{table_FoggyCULane_result}, our method consistently achieved the best performance across various scenarios. Additionally, our model achieved the highest overall F1 score of 79.85,which achieves a 0.72\% improvement over the best alternative method.

\subsubsection{Results on FoggyTusimple}As shown in the Table \ref{table_FoggyTusimple_result}, the traffic scenes in the FoggyTusimple dataset are relatively simple, resulting in smaller performance differences between methods. Nonetheless, our method achieved an F1 score of 96.95, outperforming the best existing method by 0.43\%. This further demonstrates the effectiveness of our method in foggy conditions.

\begin{figure*}[htb]
    \centering
    \includegraphics[width=\textwidth]{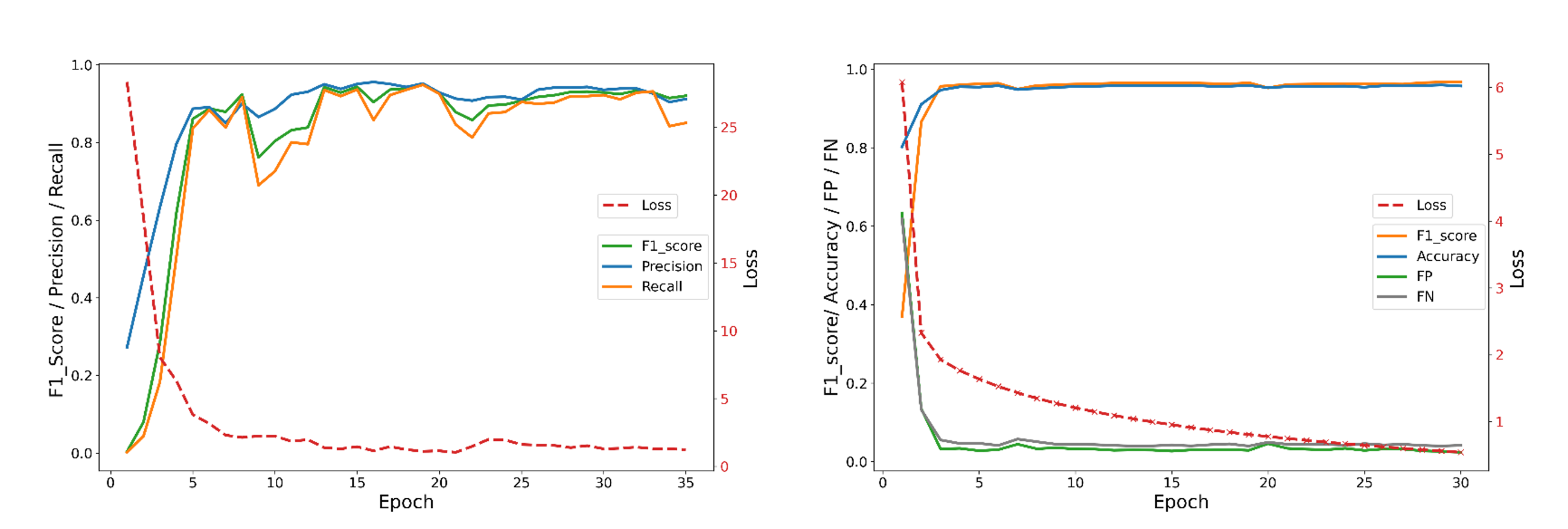}
    \caption{Several metrics of our model's performance during the training process on the FoggyLane and FoggyTusimple datasets, with results on FoggyLane on the left and FoggyTusimple on the right.}
    \label{fig_curve}
\end{figure*}

\subsection{Ablation Study}
We conducted ablation studies on the FoggyLane and FoggyCULane datasets to evaluate the effectiveness of the GFFM, DFFM, and PEM modules, using Swin-t as the backbone. As shown in Tables \ref{table_ablation_1} and \ref{table_ablation_2}, each module improves the model's performance. The GFFM module contributes the most, increasing the F1-score by 1.07 on FoggyLane and 0.41 on FoggyCULane, highlighting its ability to capture both local and global information. Adding the DFFM module further boosts the F1-score by 0.63 on FoggyLane and 0.26 on FoggyCULane. Finally, the PEM module improves edge detection, increasing the F1-score by 0.28 on FoggyLane and 0.14 on FoggyCULane.
\begin{table}[!t]
\renewcommand{\arraystretch}{1.5}
\caption{Ablation experiments of different components on the FoggyLane dataset.}
\centering
\label{table_ablation_1}
\resizebox{0.49\textwidth}{!}{
\begin{tabular}{c|ccc|c|c }
\Xhline{1pt}
Model                          & GFFM & DFFM & PEM & \textbf{F1-score} &Gflops(G) \\ \hline
Baseline        &      &       &      & 90.36                 & 13.77                   \\
                & \ding{51}    &       &      & 91.43                &11.94                    \\
                & \ding{51}    & \ding{51}     &      & 92.06                 &11.97                    \\
\textbf{PDT-Net}   & \ding{51}    & \ding{51}     & \ding{51}    & \textbf{\textcolor{blue}{92.34$\uparrow$}}        & 12.01  \\
\Xhline{1pt}
\end{tabular}
}
\end{table}

\subsection{Model deployment}
\begin{figure}
    \centering
    \includegraphics[width=0.95\columnwidth]{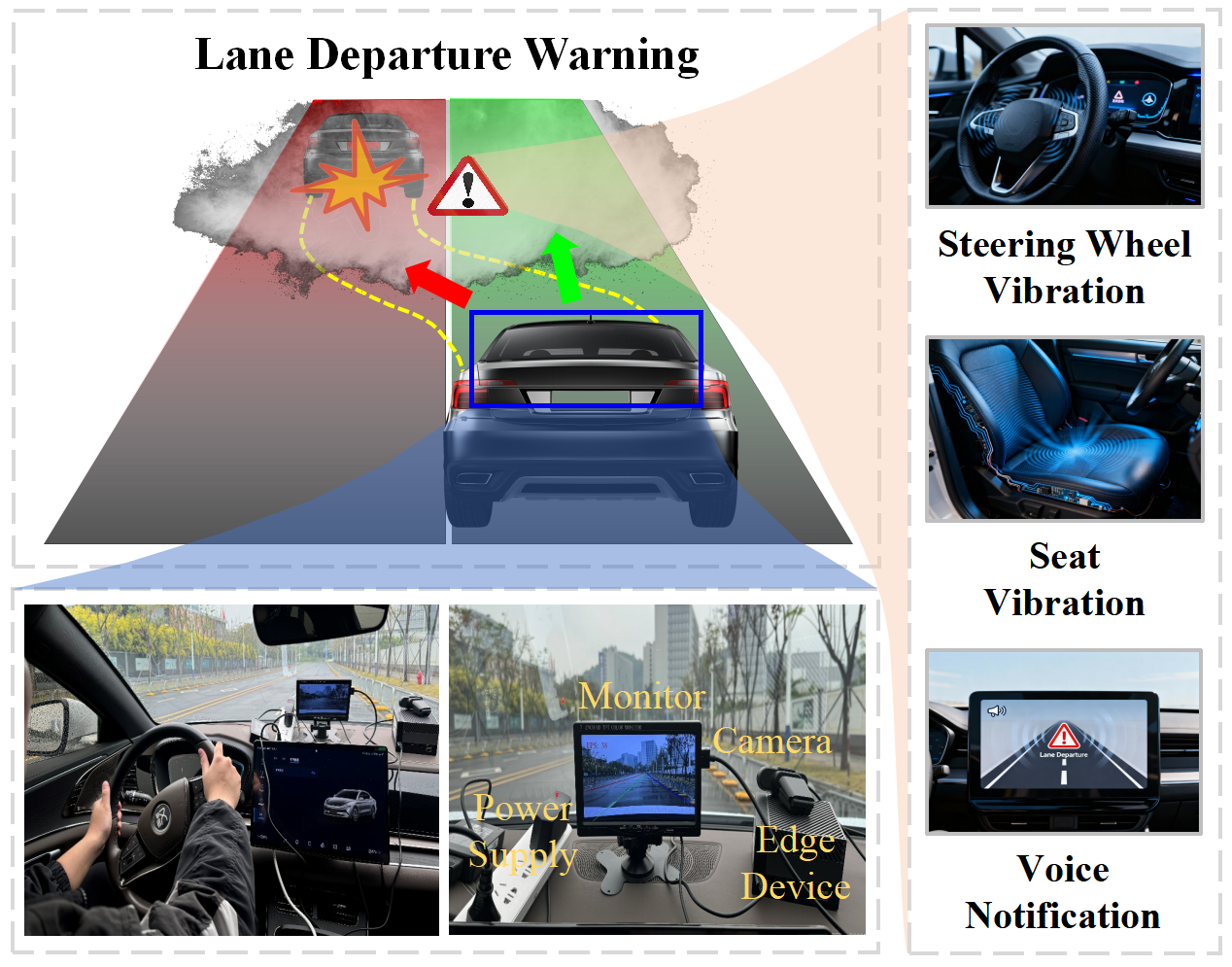}
    \caption{Lane departure warning module within the active safety warning systems framework. The module detects unintentional lane departure using a forward-facing camera and edge computing system built with NVIDIA Jetson AGX Orin for real-time lane marking recognition. Upon detecting a departure trajectory (represented by yellow dotted lines), the system triggers multi-modal warnings including steering wheel vibration, seat vibration, and voice notification with corresponding warning icons on the dashboard monitor to proactively prevent accidents.}
    \label{fig_on_car}
\end{figure}

To further evaluate the performance of our algorithm in real-world scenarios, in addition to experiments on existing datasets, we also deployed and tested the model on the vehicle. Specifically, we implemented an edge computing system on the vehicle, as illustrated in the Fig. \ref{fig_on_car}. This system comprises the NVIDIA Jetson AGX Orin edge computing device, a monitor, a camera, and a power supply unit. The results of the evaluation demonstrate that our approach can operate in real-time with high accuracy on the vehicle’s edge computing platform, despite the constraints on computational resources.
\begin{table}[htb]
\renewcommand{\arraystretch}{1.5}
\caption{Ablation experiments of different components on the FoggyCULane dataset.}
\centering
\label{table_ablation_2}
\resizebox{0.49\textwidth}{!}{
\begin{tabular}{c|ccc|c|c }
\Xhline{1pt}
Model                          & GFFM & DFFM & PEM & \textbf{F1-score} &Gflops(G) \\ \hline
Baseline        &      &       &      & 77.85                 & 13.77                   \\
                & \ding{51}    &       &      & 78.26                &11.94                    \\
                & \ding{51}    & \ding{51}     &      & 78.52                 &11.97                    \\
\textbf{PDT-Net}   & \ding{51}    & \ding{51}     & \ding{51}    & \textbf{\textcolor{blue}{78.66$\uparrow$}}        & 12.01  \\
\Xhline{1pt}
\end{tabular}
}
\end{table}
\FloatBarrier  % 禁止前面浮动体越过此线
\section{Conclusion}
This study addresses the critical challenge of lane detection in foggy conditions, which is essential for improving road safety and enhancing Advanced Driver Assistance Systems (ADAS). To tackle the issue of insufficient publicly available datasets for foggy lane detection, we constructed the FoggyLane dataset, a real-world dataset specifically targeting lane detection in foggy environments, and synthesized two additional datasets, FoggyCULane and FoggyTusimple, to fill the gap in fog-specific data.

In response to the difficulties posed by foggy scenarios, we propose a robust Prior-Guided Dynamic Feature Fusion Transformer framework for real-time lane detection. This method incorporates a Global Feature Fusion Module to capture the relationship between local and global features in foggy images, a Dynamic Feature Fusion Module to model the structural and positional relationships of lane instances, and a Prior-Guided Edge Enhancement Module to recover lost edge details. Extensive experiments show that our approach significantly improves lane detection accuracy compared to other advanced methods, both in clear and foggy conditions, with F1-scores of 95.04\%, 79.85\%, and 96.95\% on the FoggyLane, FoggyCULane, and FoggyTusimple datasets, respectively.

Moreover, with TensorRT acceleration, our method achieves an inference speed of 38.4 FPS on the NVIDIA Jetson AGX Orin, confirming its practical applicability in real-world scenarios. By improving the precision of lane detection, our framework contributes to active safety warning systems, helping prevent accidents in foggy conditions and enhancing road safety, particularly in challenging driving environments.

\section*{Acknowledgement}
This project is jointly supported by National Natural Science Foundation of China (Nos. 52172350, W2421069 and 51775565), the Guangdong Basic and Applied Research Foundation (No. 2022B1515120072), the Guangzhou Science and Technology Plan Project (No. 2024B01W0079), the Science and Technology Planning Project of Guangdong Province (No. 2023B1212060029).\textbf{\textcolor{blue}{We would like to release our source code and dataset.}}

\section{Data Availability}

Data will be made available on request.

\section{Declaration of Competing Interest}

The authors declare that they have no known competing financial interests or personal relationships that could have appeared to influence the work reported in this article.

\section{Acknowledgement}

This project is jointly supported by National Natural Science Foundation of China (Nos.52172350, 51775565), Guangdong Basic and Applied Research Foundation (Nos.2\\021B1515120032, 2022B1515120072), Guangzhou Science and Technology Plan Project (Nos.2024B01W0079,20\\2206030005), Science and Technology Planning Project of Guangdong Province (No.2023B1212060029).


\printcredits

\bibliographystyle{cas-model2-names}
\bibliography{cas-refs}

@article{son2015real,
  title={Real-time illumination invariant lane detection for lane departure warning system},
  author={Son, Jongin and Yoo, Hunjae and Kim, Sanghoon and Sohn, Kwanghoon},
  journal={Expert Systems with Applications},
  volume={42},
  number={4},
  pages={1816--1824},
  year={2015},
  publisher={Elsevier},
  doi={https://doi.org/10.1016/j.eswa.2014.10.024}
}

@inproceedings{hou2016efficient,
  title={An efficient lane markings detection and tracking method based on vanishing point constraints},
  author={Hou, Changzheng and Hou, Jin and Yu, Chaochao},
  booktitle={Chinese Control Conference (CCC)},
  pages={6999--7004},
  year={2016},
  organization={IEEE},
  doi={10.1109/ChiCC.2016.7554460}
}

@inproceedings{neven2018towards,
  title={Towards end-to-end lane detection: an instance segmentation approach},
  author={Neven, Davy and De Brabandere, Bert and Georgoulis, Stamatios and Proesmans, Marc and Van Gool, Luc},
  booktitle={2018 IEEE Intelligent Vehicles Symposium (IV)},
  pages={286--291},
  year={2018},
  organization={IEEE},
  doi={10.1109/IVS.2018.8500547}
}

@article{abualsaud2021laneaf,
  title={LaneAF: Robust multi-lane detection with affinity fields},
  author={Abualsaud, Hala and Liu, Sean and Lu, David B and Situ, Kenny and Rangesh, Akshay and Trivedi, Mohan M},
  journal={IEEE Robotics and Automation Letters},
  volume={6},
  number={4},
  pages={7477--7484},
  year={2021},
  publisher={IEEE},
  doi={10.1109/LRA.2021.3098066}
}

@inproceedings{chen2019pointlanenet,
  title={Pointlanenet: Efficient end-to-end cnns for accurate real-time lane detection},
  author={Chen, Zhenpeng and Liu, Qianfei and Lian, Chenfan},
  booktitle={2019 IEEE Intelligent Vehicles Symposium (IV)},
  pages={2563--2568},
  year={2019},
  doi={10.1109/IVS.2019.8813778}
}

@inproceedings{yu2018deep,
  title={Deep layer aggregation},
  author={Yu, Fisher and Wang, Dequan and Shelhamer, Evan and Darrell, Trevor},
  booktitle={Proceedings of the IEEE Conference on Computer Vision and Pattern Recognition},
  pages={2403--2412},
  year={2018},
  doi={10.1109/CVPR.2018.00255}
}

@inproceedings{Caltech-Lanes,
  title={Real time detection of lane markers in urban streets},
  author={Aly, Mohamed},
  booktitle={2008 IEEE Intelligent Vehicles Symposium (IV)},
  pages={7--12},
  year={2008},
  doi={10.1109/IVS.2008.4621152}
}

@inproceedings{redmon2016you,
  title={You only look once: Unified, real-time object detection},
  author={Redmon, J},
  booktitle={Proceedings of the IEEE Conference on Computer Vision and Pattern Recognition},
  pages={779--788},
  year={2016},
  doi={10.1109/CVPR.2016.91}
}

@article{farhadi2018yolov3,
  title={Yolov3: An incremental improvement},
  author={Farhadi, Ali and Redmon, Joseph},
  journal={arXiv preprint arXiv:1804.02767},
  pages={1--6},
  year={2018},
  doi={10.48550/arXiv.1804.02767}
}

@article{huang2020dsnet,
  title={DSNet: Joint semantic learning for object detection in inclement weather conditions},
  author={Huang, Shih-Chia and Le, Trung-Hieu and Jaw, Da-Wei},
  journal={IEEE Transactions on Pattern Analysis and Machine Intelligence},
  volume={43},
  number={8},
  pages={2623--2633},
  year={2020},
  publisher={IEEE},
  doi={10.1109/TPAMI.2020.2977911}
}

@inproceedings{chen2018domain,
  title={Domain adaptive faster r-cnn for object detection in the wild},
  author={Chen, Yuhua and Li, Wen and Sakaridis, Christos and Dai, Dengxin and Van Gool, Luc},
  booktitle={Proceedings of the IEEE Conference on Computer Vision and Pattern Recognition},
  pages={3339--3348},
  year={2018},
  doi={10.1109/CVPR.2018.00352}
}

@inproceedings{zhang2021domain,
  title={Domain adaptive yolo for one-stage cross-domain detection},
  author={Zhang, Shizhao and Tuo, Hongya and Hu, Jian and Jing, Zhongliang},
  booktitle={Asian Conference on Machine Learning},
  pages={785--797},
  year={2021},
  organization={PMLR}
}

@inproceedings{hnewa2021multiscale,
  title={Multiscale domain adaptive yolo for cross-domain object detection},
  author={Hnewa, Mazin and Radha, Hayder},
  booktitle={Proceedings of the IEEE International Conference on Image Processing},
  pages={3323--3327},
  year={2021},
  doi={10.1109/ICIP42928.2021.9506039}
}

@inproceedings{li2023domain,
  title={Domain adaptive object detection for autonomous driving under foggy weather},
  author={Li, Jinlong and Xu, Runsheng and Ma, Jin and Zou, Qin and Ma, Jiaqi and Yu, Hongkai},
  booktitle={Proceedings of the IEEE/CVF Winter Conference on Applications of Computer Vision},
  pages={612--622},
  year={2023}
}

@ARTICLE{9052469,
  author={Wang, Jingdong and Sun, Ke and Cheng, Tianheng and Jiang, Borui and Deng, Chaorui and Zhao, Yang and Liu, Dong and Mu, Yadong and Tan, Mingkui and Wang, Xinggang and Liu, Wenyu and Xiao, Bin},
  journal={IEEE Transactions on Pattern Analysis and Machine Intelligence}, 
  title={Deep High-Resolution Representation Learning for Visual Recognition}, 
  year={2021},
  volume={43},
  number={10},
  pages={3349-3364},
  doi={10.1109/TPAMI.2020.2983686}
}

@misc{Tusimple,
  author = {Tusimple},
  howpublished = {\url{https://github.com/TuSimple/ tusimple-benchmark}},
  year = {2017}
}

@misc{YouTube,
  author = {YouTube},
  howpublished = {\url{https://www.youtube.com/}},
  year = {Accessed: 2024}
}

@article{zhang2024robust,
  title={A robust and real-time lane detection method in low-light scenarios to advanced driver assistance systems},
  author={Zhang, Ronghui and Peng, Jingtao and Gou, Wanting and Ma, Yuhang and Chen, Junzhou and Hu, Hongyu and Li, Weihua and Yin, Guodong and Li, Zhiwu},
  journal={Expert Systems with Applications},
  volume={256},
  pages={124923},
  year={2024},
  publisher={Elsevier},
  doi={10.1016/j.eswa.2024.124923}
}

@inproceedings{LLAMAS,
  title={Unsupervised labeled lane markers using maps},
  author={Behrendt, Karsten and Soussan, Ryan},
  booktitle={Proceedings of the IEEE/CVF International Conference on Computer Vision Workshops},
  pages={832-839},
  year={2019},
  doi={10.1109/ICCVW.2019.00111}
}

@inproceedings{Apolloscape,
  title={The apolloscape dataset for autonomous driving},
  author={Huang, Xinyu and Cheng, Xinjing and Geng, Qichuan and Cao, Binbin and Zhou, Dingfu and Wang, Peng and Lin, Yuanqing and Yang, Ruigang},
  booktitle={Proceedings of the IEEE Conference on Computer Vision and Pattern Recognition Workshops},
  pages={954--960},
  year={2018},
  doi={10.1109/CVPRW.2018.00141}
}

@inproceedings{yu2020bdd100k,
  title={Bdd100k: A diverse driving dataset for heterogeneous multitask learning},
  author={Yu, Fisher and Chen, Haofeng and Wang, Xin and Xian, Wenqi and Chen, Yingying and Liu, Fangchen and Madhavan, Vashisht and Darrell, Trevor},
  booktitle={Proceedings of the IEEE/CVF Conference on Computer Vision and Pattern Recognition},
  pages={2636--2645},
  year={2020},
  doi={10.1109/CVPR42600.2020.00271}
}

@inproceedings{pan2018spatial,
  title={Spatial as deep: Spatial cnn for traffic scene understanding},
  author={Pan, Xingang and Shi, Jianping and Luo, Ping and Wang, Xiaogang and Tang, Xiaoou},
  booktitle={Proceedings of the AAAI Conference on Artificial Intelligence},
  volume={32},
  number={1},
  year={2018}
}

@inproceedings{xu2020curvelane,
  title={Curvelane-nas: Unifying lane-sensitive architecture search and adaptive point blending},
  author={Xu, Hang and Wang, Shaoju and Cai, Xinyue and Zhang, Wei and Liang, Xiaodan and Li, Zhenguo},
  booktitle={European Conference on Computer Vision},
  pages={689--704},
  year={2020},
  organization={Springer},
  doi={10.1007/978-3-030-58555-6_41}
}

@ARTICLE{he2011darkchannel,
  author={He, Kaiming and Sun, Jian and Tang, Xiaoou},
  journal={IEEE Transactions on Pattern Analysis and Machine Intelligence}, 
  title={Single Image Haze Removal Using Dark Channel Prior}, 
  year={2011},
  volume={33},
  number={12},
  pages={2341-2353},
  doi={10.1109/TPAMI.2010.168}
}

@InProceedings{chen2021S2R,
    author={Chen, Xiaotian and Wang, Yuwang and Chen, Xuejin and Zeng, Wenjun},
    title={S2R-DepthNet: Learning a Generalizable Depth-Specific Structural Representation},
    booktitle={Proceedings of the IEEE/CVF Conference on Computer Vision and Pattern Recognition },
    month= {June},
    year= {2021},
    pages= {3034-3043},
    doi={10.1109/CVPR46437.2021.00305}
}

@article{harald1924theorie,
  title={Theorie der horizontalen sichtweite: Kontrast und sichtweite},
  author={Harald, Koschmieder},
  journal={Keim and Nemnich, Munich},
  volume={12},
  pages={33–53},
  year={1924},
  doi={10.1007/978-3-663-04661-5_2}
}

@article{nie2022sensor,
  title={Foggy lane dataset synthesized from monocular images for lane detection algorithms},
  author={Nie, Xiangyu and Xu, Zhejun and Zhang, Wei and Dong, Xue and Liu, Ning and Chen, Yuanfeng},
  journal={Sensors},
  volume={22},
  number={14},
  pages={5210},
  year={2022},
  publisher={MDPI},
  doi={10.3390/s22145210}
}

@inproceedings{zheng2021resa,
  title={Resa: Recurrent feature-shift aggregator for lane detection},
  author={Zheng, Tu and Fang, Hao and Zhang, Yi and Tang, Wenjian and Yang, Zheng and Liu, Haifeng and Cai, Deng},
  booktitle={Proceedings of the AAAI Conference on Artificial Intelligence},
  volume={35},
  number={4},
  pages={3547--3554},
  year={2021},
  doi={10.1609/aaai.v35i4.16469}
}

@article{li2019line,
  title={Line-cnn: End-to-end traffic line detection with line proposal unit},
  author={Li, Xiang and Li, Jun and Hu, Xiaolin and Yang, Jian},
  journal={IEEE Transactions on Intelligent Transportation Systems},
  volume={21},
  number={1},
  pages={248--258},
  year={2019},
  publisher={IEEE},
  doi={10.1109/TITS.2019.2890870}
}

@inproceedings{tabelini2021keep,
  title={Keep your eyes on the lane: Real-time attention-guided lane detection},
  author={Tabelini, Lucas and Berriel, Rodrigo and Paixao, Thiago M and Badue, Claudine and De Souza, Alberto F and Oliveira-Santos, Thiago},
  booktitle={Proceedings of the IEEE/CVF Conference on Computer Vision and Pattern Recognition},
  pages={294--302},
  year={2021},
  doi={10.1109/CVPR46437.2021.00036}
}

@inproceedings{zheng2022clrnet,
  title={Clrnet: Cross layer refinement network for lane detection},
  author={Zheng, Tu and Huang, Yifei and Liu, Yang and Tang, Wenjian and Yang, Zheng and Cai, Deng and He, Xiaofei},
  booktitle={Proceedings of the IEEE/CVF Conference on Computer Vision and Pattern Recognition},
  pages={898--907},
  year={2022},
  doi={10.1109/CVPR52688.2022.00097}
}

@article{ran2023flamnet,
  title={Flamnet: A flexible line anchor mechanism network for lane detection},
  author={Ran, Hao and Yin, Yunfei and Huang, Faliang and Bao, Xianjian},
  journal={IEEE Transactions on Intelligent Transportation Systems},
  Volume={24},
  pages={12767--12778},
  year={2023},
  publisher={IEEE},
  doi={doi: 10.1109/TITS.2023.3290991}
}

@inproceedings{tabelini2021polylanenet,
  title={Polylanenet: Lane estimation via deep polynomial regression},
  author={Tabelini, Lucas and Berriel, Rodrigo and Paixao, Thiago M and Badue, Claudine and De Souza, Alberto F and Oliveira-Santos, Thiago},
  booktitle={Proceedings of the IEEE International Conference on Pattern Recognition},
  pages={6150--6156},
  year={2021},
  organization={IEEE},
  doi={10.1109/ICPR48806.2021.9412265}
}

@inproceedings{liu2021end,
  title={End-to-end lane shape prediction with transformers},
  author={Liu, Ruijin and Yuan, Zejian and Liu, Tie and Xiong, Zhiliang},
  booktitle={Proceedings of the IEEE/CVF Winter Conference on Applications of Computer Vision},
  pages={3694--3702},
  year={2021},
  doi={10.1109/WACV48630.2021.00374}
}

@article{dai2024dbnet,
  title={DBNet: A Curve-based Dynamic Association Framework for Lane Detection},
  author={Dai, Xinguang and Xie, Jun and Zhang, Guoxin and Chang, Kenglun and Chen, Feng and Wang, Zhepeng and Tang, Chunming},
  journal={IEEE Transactions on Intelligent Vehicles},
  year={2024},
  publisher={IEEE},
  doi={10.1109/TIV.2024.3370213}
}

@inproceedings{yoo2020end,
  title={End-to-end lane marker detection via row-wise classification},
  author={Yoo, Seungwoo and Lee, Hee Seok and Myeong, Heesoo and Yun, Sungrack and Park, Hyoungwoo and Cho, Janghoon and Kim, Duck Hoon},
  booktitle={Proceedings of the IEEE/CVF Conference on Computer Vision and Pattern Recognition Workshops},
  pages={1006--1007},
  year={2020},
  doi={10.1109/CVPRW50498.2020.00511}
}

@inproceedings{qin2020ultra,
  title={Ultra fast structure-aware deep lane detection},
  author={Qin, Zequn and Wang, Huanyu and Li, Xi},
  booktitle={European Conference on Computer Vision},
  pages={276--291},
  year={2020},
  organization={Springer},
  doi={10.1007/978-3-030-58586-0_17}
}

@inproceedings{liu2021condlanenet,
  title={Condlanenet: a top-to-down lane detection framework based on conditional convolution},
  author={Liu, Lizhe and Chen, Xiaohao and Zhu, Siyu and Tan, Ping},
  booktitle={Proceedings of the IEEE/CVF International Conference on Computer Vision},
  pages={3773--3782},
  year={2021},
  doi={10.1109/ICCV48922.2021.00375}
}

@inproceedings{liu2021swin,
  title={Swin transformer: Hierarchical vision transformer using shifted windows},
  author={Liu, Ze and Lin, Yutong and Cao, Yue and Hu, Han and Wei, Yixuan and Zhang, Zheng and Lin, Stephen and Guo, Baining},
  booktitle={Proceedings of the IEEE/CVF International Conference on Computer Vision},
  pages={10012--10022},
  year={2021},
  doi={10.1109/ICCV48922.2021.00986}
}

@ARTICLE{tian2020conditional,
  author={Tian, Zhi and Zhang, Bowen and Chen, Hao and Shen, Chunhua},
  journal={IEEE Transactions on Pattern Analysis and Machine Intelligence}, 
  title={Instance and Panoptic Segmentation Using Conditional Convolutions}, 
  year={2023},
  volume={45},
  number={1},
  pages={669-680},
  doi={10.1109/TPAMI.2022.3145407}
}

@inproceedings{vit,
  title={An Image is Worth 16x16 Words: Transformers for Image Recognition at Scale},
  author={Dosovitskiy, Alexey and Beyer, Lucas and Kolesnikov, Alexander and Weissenborn, Dirk and Zhai, Xiaohua and Unterthiner, Thomas and Dehghani, Mostafa and Minderer, Matthias and Heigold, Georg and Gelly, Sylvain and others},
  booktitle={International Conference on Learning Representations},
  pages={1--17},
  year={2021},
}

@misc{googlemap,
  author = {GoogleMaps},
  howpublished = {\url{https://www.google.com/maps}},
  year = {Accessed: 2024}
}

@misc{Pexels,
  author = {Pexels},
  howpublished = {\url{https://www.pexels.com/zh-cn/}},
  year = {2025}
}

@ARTICLE{tits3,
  author={Zhao, Jing and Qiu, Zengyu and Hu, Huiqin and Sun, Shiliang},
  journal={IEEE Transactions on Intelligent Transportation Systems}, 
  title={HWLane: HW-Transformer for Lane Detection}, 
  year={2024},
  volume={25},
  number={8},
  pages={9321-9331},
  doi={10.1109/TITS.2024.3386531}}

@ARTICLE{tits4,
  author={Wang, Shengqi and Liu, Junmin and Cao, Xiangyong and Song, Zengjie and Sun, Kai},
  journal={IEEE Transactions on Intelligent Transportation Systems}, 
  title={Polar R-CNN: End-to-End Lane Detection With Fewer Anchors}, 
  year={2025},
  pages={1-13},
  doi={10.1109/TITS.2025.3564979}}

@techreport{NHTSA2025,
  author = {{National Highway Traffic Safety Administration}},
  title = {{Traffic Safety Facts 2023 Data: Rural/Urban Traffic Fatalities}},
  institution = {{U.S. Department of Transportation}},
  year = {2025},
  month = {June},
  number = {DOT HS 813 728},
  type = {Data Report},
  url = {https://crashstats.nhtsa.dot.gov/Api/Public/ViewPublication/813728},
  note = {Accessed: 2025}
}

@article{AAP1,
title = {Ensuring SOTIF: Enhanced object detection techniques for autonomous driving},
journal = {Accident Analysis \& Prevention},
volume = {218},
pages = {108094},
year = {2025},
doi = {https://doi.org/10.1016/j.aap.2025.108094},
author = {Sifen Wang and Zhangyu Wang and Sheng Hong and Pengcheng Wang and Shaowei Zhang},
}

@article{AAP2,
title = {Crash report data analysis for creating scenario-wise, spatio-temporal attention guidance to support computer vision-based perception of fatal crash risks},
journal = {Accident Analysis \& Prevention},
volume = {151},
pages = {105962},
year = {2021},
doi = {https://doi.org/10.1016/j.aap.2020.105962},
author = {Yu Li and Muhammad Monjurul Karim and Ruwen Qin and Zeyi Sun and Zuhui Wang and Zhaozheng Yin},
}

@article{AAP3,
title = {Real-time accident anticipation for autonomous driving through monocular depth-enhanced 3D modeling},
journal = {Accident Analysis \& Prevention},
volume = {207},
pages = {107760},
year = {2024},
doi = {https://doi.org/10.1016/j.aap.2024.107760},
author = {Haicheng Liao and Yongkang Li and Zhenning Li and Zilin Bian and Jaeyoung Lee and Zhiyong Cui and Guohui Zhang and Chengzhong Xu},
}

@article{AAP4,
title = {Advances and applications of computer vision techniques in vehicle trajectory generation and surrogate traffic safety indicators},
journal = {Accident Analysis \& Prevention},
volume = {191},
pages = {107191},
year = {2023},
doi = {https://doi.org/10.1016/j.aap.2023.107191},
author = {Mohamed Abdel-Aty and Zijin Wang and Ou Zheng and Amr Abdelraouf},
}

@article{AAP5,
title = {Automated vehicle data pipeline for accident reconstruction: New insights from LiDAR, camera, and radar data},
journal = {Accident Analysis \& Prevention},
volume = {180},
pages = {106923},
year = {2023},
doi = {https://doi.org/10.1016/j.aap.2022.106923},
author = {Joe Beck and Ramin Arvin and Steve Lee and Asad Khattak and Subhadeep Chakraborty},
}

@article{AAP6,
title = {Integrating visual and community environments in a motorcycle crash and casualty estimation},
journal = {Accident Analysis \& Prevention},
volume = {208},
pages = {107792},
year = {2024},
doi = {https://doi.org/10.1016/j.aap.2024.107792},
author = {Yujin Kim and Hwasoo Yeo and Lisa Lim and Byeongjoon Noh},
}

@article{AAP7,
title = {Effect of warning message on driver’s stop/go decision and red-light-running behaviors under fog condition},
journal = {Accident Analysis \& Prevention},
volume = {150},
pages = {105906},
year = {2021},
issn = {0001-4575},
doi = {https://doi.org/10.1016/j.aap.2020.105906},
author = {Yuting Zhang and Xuedong Yan and Xiaomeng Li},
}

@article{AAP8,
title = {Driving risks assessment and in-vehicle warning design for improving work zone safety},
journal = {Accident Analysis \& Prevention},
volume = {215},
pages = {107991},
year = {2025},
issn = {0001-4575},
doi = {https://doi.org/10.1016/j.aap.2025.107991},
author = {Junyu Hang and Xiaomeng Li and Xuedong Yan and Ke Duan and Qingchun Wang and Qingwan Xue},
}

\end{document}